\documentclass[review]{elsarticle}

\usepackage{lineno}
\usepackage[hidelinks]{hyperref}
\usepackage{graphicx}
\usepackage{caption}
\usepackage{tabularx}
	\newcolumntype{L}{>{\raggedright\arraybackslash}X}
\modulolinenumbers[5]
\usepackage{url}
\usepackage{enumitem}
\usepackage{algpseudocode}
\usepackage{pgfplots}
\usepackage{pgfplotstable}
\usepackage{amsmath}
\usepackage{booktabs}
\usepackage{cellspace}
\usepackage{siunitx}
\usepackage{setspace}
\usepackage[margin=1.5in]{geometry}
\usepackage{amsmath,amssymb,amsfonts, picins}
\usepackage{algorithm}
\usepackage[utf8]{inputenc}

\newcommand{\squishlist}{
 \begin{list}{$\bullet$}
  { \setlength{\itemsep}{0pt}
     \setlength{\parsep}{3pt}
     \setlength{\topsep}{3pt}
     \setlength{\partopsep}{0pt}
     \setlength{\leftmargin}{1.5em}
     \setlength{\labelwidth}{1em}
     \setlength{\labelsep}{0.5em}}
     }

\newcommand{\squishlisttwo}{
 \begin{list}{$\bullet$}
  { \setlength{\itemsep}{0pt}
     \setlength{\parsep}{0pt}
    \setlength{\topsep}{0pt}
    \setlength{\partopsep}{0pt}
    \setlength{\leftmargin}{2em}
    \setlength{\labelwidth}{1.5em}
    \setlength{\labelsep}{0.5em} } }

\newcommand{\squishend}{
\end{list}  }

\usepackage{comment}

\newcounter{phase}[algorithm]
\newlength{\phaserulewidth}
\newcommand{\setphaserulewidth}{\setlength{\phaserulewidth}}
\newcommand{\phase}[1]{%
  \vspace{-1.25ex}
  \Statex\leavevmode\llap{\rule{\dimexpr\labelwidth+\labelsep}{\phaserulewidth}}\rule{\linewidth}{\phaserulewidth}
  \Statex\strut\refstepcounter{phase}\textit{Phase~\thephase~--~#1}
  \vspace{-1.25ex}\Statex\leavevmode\llap{\rule{\dimexpr\labelwidth+\labelsep}{\phaserulewidth}}\rule{\linewidth}{\phaserulewidth}}
\makeatother

\setphaserulewidth{.7pt}

\newenvironment{algocolor}{%
   \setlength{\parindent}{0pt}
   \itshape
   \color{black}
}{}

\journal{Pattern Recognition} 

\begin{document}

\begin{frontmatter}

\title{A Multi-Modal Transformer Network for Action Detection}


\author[mymainaddress]{Matthew Korban}
\ead{acw6ze@virginia.edu}

\author[mysecondaryaddress]{Peter Youngs}
\ead{pay2n@virginia.edu}

\author[mymainaddress]{Scott T. Acton\corref{mycorrespondingauthor}}
\cortext[mycorrespondingauthor]{Corresponding author}
\ead{acton@virginia.edu}

\address[mymainaddress]{Department of Electrical and Computer Engineering, University of Virginia, Charlottesville, VA 22904}
\address[mysecondaryaddress]{Department of Curriculum, Instruction and Special Education, University of Virginia, VA 22904}

\begin{abstract}
This paper proposes a novel multi-modal transformer network for detecting actions in untrimmed videos. To enrich the action features, our transformer network utilizes a new multi-modal attention mechanism that computes the correlations between different spatial and motion modalities combinations. Exploring such correlations for actions has not been attempted previously. To use the motion and spatial modality more effectively, we suggest an algorithm that corrects the motion distortion caused by camera movement. Such motion distortion, common in untrimmed videos, severely reduces the expressive power of motion features such as optical flow fields. 
Our proposed algorithm outperforms the state-of-the-art methods on two public benchmarks, THUMOS14 and ActivityNet.
We also conducted comparative experiments on our new instructional activity dataset, including a large set of challenging classroom videos captured from elementary schools.

\end{abstract}


\end{frontmatter}

\nolinenumbers

\section{Introduction}
\label{sec:intro}
Action detection is temporally localizing action class instances in untrimmed videos. Action sequences are represented by spatial and temporal components, which jointly define the meaning of various actions. For example, the action ``throwing a ball" is characterized by its spatial components, the image pixels of the ball, and its movement during the action sequence.
A popular way to represent such spatial and temporal components of actions are \emph{RGB} images and \emph{optical flows}, respectively \cite{ali2008PAMI}.  
However, action detection using both RGB and optical flow modalities is challenging. The two main challenges are separated RGB and optical flow modalities, and camera movement \cite{yin2018CVPR}. We will discuss the specific challenges above and our solutions to handle them in the following.

\subsection{Multi-modal attention for solving separated RGB and optical flow}
\label{Sec:intro_MMA}

\begin{figure}[!htbp]
	\centering 
		\begin{tabular}{c}
              \includegraphics[height=0.36\textheight]{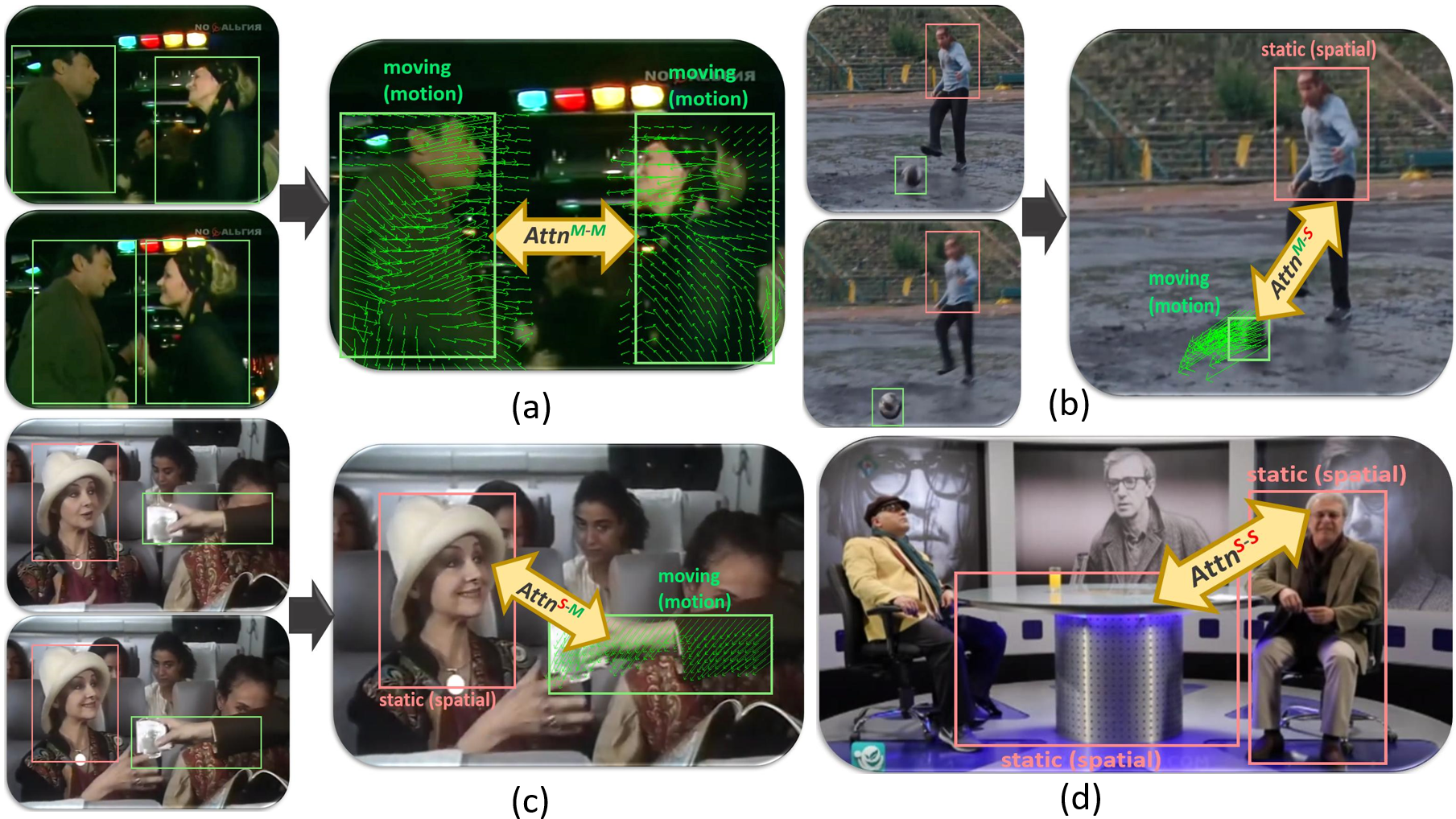}
    	\end{tabular}
	\caption{
 \textcolor{black}{(a) motion-to-motion attention ($Attn^{M-M}$): between both moving subjects/objects of interest (SOI) for the action ``dancing'' where the two persons get toward each other.
 The motion vectors for the SOIs (in green) in the right large images are calculated based on the two consecutive left (smaller) frames. 
(b) motion-to-spatial attention ($Attn^{M-S}$): between moving SOI, such as a ball, and static ones (the person). 
(c) spatial-to-motion attention ($Attn^{S-M}$): between a static person and a moving hand, holding glasses in the action ``offering a drink''. 
(d) spatial-to-spatial attention ($Attn^{S-S}$): all the SOI (person and table) are static such as ``sitting at a table''.
The raw images are collected from an in-the-wild activity dataset \cite{gu2018CVPR}.}
  ~\label{Fig:att_examples}}
\end{figure}

Many actions are defined by both moving and static objects/subjects. To model such spatial-temporal information in actions, many researchers have used optical flow fields and RGB images to represent the motion of actions and their spatial properties, respectively \cite{vahdani2022PAMI}. Optical flow is a powerful motion modality that is computed from RGB images \cite{kong2021ICRA}. However, such a computation is complex and typically iterative and non-differentiable, which cannot be achieved easily within the context of an action detection network whose loss function is designed explicitly to model activities. Moreover, the current well-known action modeling baselines such as I3D \cite{carreira2017CVPR} provide optical flow data separately from RGB images. Hence, the exiting action detection algorithms either used RGB images and optical flow fields independently in two streams or just combined them without differentiating these modalities \cite{vahdani2022PAMI}. Many actions, however, are defined not only within the spatial and motion domains but also based on the interactions between these domains. Some examples are shown in Fig. \ref{Fig:att_examples}. Computing such interactions is, however, challenging as two spatial and motion domains are represented in different modalities. 

Some methods have evaluated several aspects of optical flow fields in action recognition and detection. \cite{guo2010ICAVSS} suggested that some features such as velocity, gradient, and divergence represented in optical flow fields are effective in action recognition.  \cite{sevilla2018GCPR} investigated the correlations between optical flow and action recognition accuracy based on several well-known optical flow estimation methods.  \cite{sun2018CVPR} proposed a deep network that extracts effective optical flow for action recognition. However, there has not been any effective approach to capture the correlations between optical flow and RGB images. Such correlations, as stated above, are important in modeling many actions though they cannot be inferred easily by standard action detection algorithms. So, we propose an effective strategy to capture the correlations between optical flow and RGB images using our multi-modal attention mechanism. Our variations of multi-modal attention are shown in Fig. \ref{Fig:att_examples}, and more details about this figure are in Section \ref{Sec:MMA}. 


\subsection{Motion distortion correction for solving camera movement}
\begin{figure}[!htbp]
	\centering 
		\begin{tabular}{c}
		\includegraphics[height=0.45\textheight]{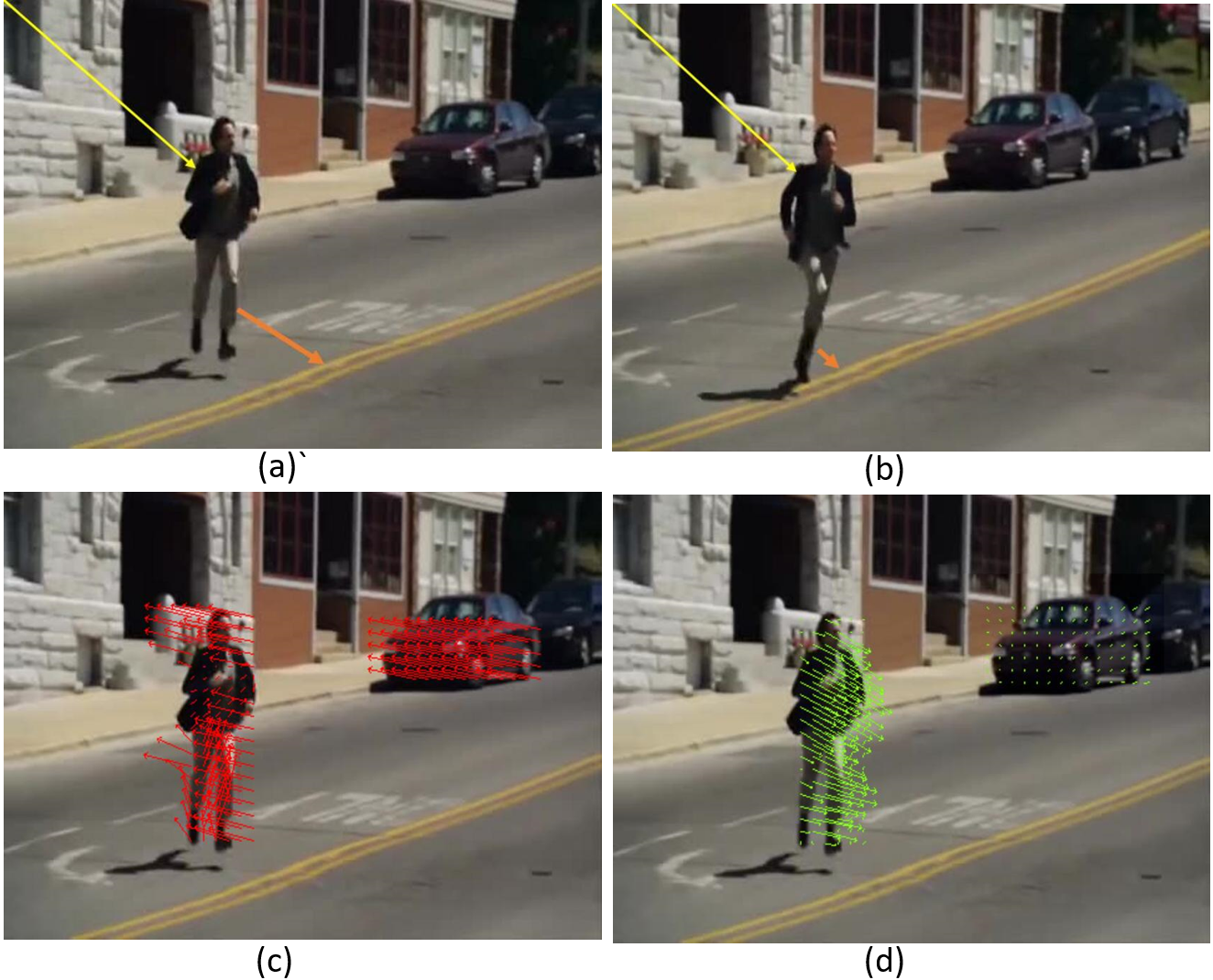} 
    	\end{tabular}
	\caption{Illustration of the results of our motion distortion correction algorithm: (a) and (b) show two consecutive sampled frames of an action sequence collected from an in-the-wild activity dataset \cite{gu2018CVPR}.  In this sequence, the person runs toward the image frame's lower part. Due to the spatial-temporal inconsistency caused by the camera movement, the actual motion is distorted. Specifically, in reality, the person is getting closer to a reference orange street line (shown as orange vectors), which is inconsistent with the person's spatial location with respect to the image origin (yellow vectors). (c) shows the optical flow motion vectors corresponding to the person and the parked car obtained from a standard state-of-the-art algorithm \cite{kong2021ICRA}. Due to the motion distortion caused by the camera movement, the motion vectors for the person and stationary parked car are incorrect. (d) shows our corrected optical flow motion vectors for the person and the stationary car. ~\label{Fig:motion_distort}}
\end{figure}

Untrimmed action videos are often captured in the wild, where camera movement is common. Such camera movement can significantly distort the motion represented by optical flow fields as it causes spatial-temporal inconsistency. An example is shown in Fig. \ref{Fig:motion_distort}, where the person is running toward the southwest and getting closer to the orange street line in two consecutive frames (Fig. \ref{Fig:motion_distort} (a) and (b)) and the car is stationary. A standard state-of-the-art optical flow estimation algorithm such as \cite{kong2021ICRA}  fails to extract the correct optical flow motion vectors (Fig. \ref{Fig:motion_distort} (c)) because of the spatial-temporal inconsistency caused by the camera movement. In this example, the person's movement (orange arrows) is inconsistent with respect to his spatial location in the image (yellow arrows). To solve this issue, we propose a motion distortion correction algorithm with corrected results for the moving person and the stationary car shown in Fig. \ref{Fig:motion_distort} (d). We show more examples of our motion distortion correction algorithm tested on a sample of THUMOS14 \cite{idrees2017thumos} 
 later in Fig. \ref{Fig:exam_thu}.

Some approaches tried to include the camera movement factor in the optical flow extraction process.  \cite{yin2018CVPR} estimated the camera pose jointly with optical flow and depth maps using a complex network architecture. Similarly, \cite{ranjan2019CVPR} proposed that a collaboration between camera pose, optical flow, and depth map estimation is useful. However, these methods have some issues: (1) such methods require complicated design and training; (2) these approaches have not been validated for their robustness under camera movements and in real-world applications such as temporal action detection to show the practical reliability of the extracted optical flows. We, however, propose a simple yet effective approach that does not require training to improve the optical flow extraction. Furthermore, we validated our improved optical flows in a real-world scenario (action detection) and on several public benchmarks where camera movement is common.

The main \textbf{contributions} of this paper are summarized as follows:

\begin{itemize}
    \item We propose a novel transformer network for action detection, including a new multi-modal attention mechanism to effectively capture spatial-temporal features from RGB and optical flow modalities.
    \item We introduce a new motion distortion correction algorithm to handle camera movements in the wild that does not require any training, and its reliability is validated in a real-world application (action detection).
    \item We conducted new experiments on our newly collected dataset depicting instructional activities within elementary school settings.
    \item Our proposed method outperforms the state-of-the-art approaches on two public benchmarks as well as on our collected dataset.
 \end{itemize}

\section{Related work}
\label{sec:RW}
Two main related topics to our work are ``action recognition'' and ``temporal action detection''. Action recognition aims to classify trimmed and often short action sequences \cite{herath2017IVC, wang2013ICCV}. In contrast, action detection intents to identify action instances in untrimmed and usually long videos \cite{zhao2017ICCV, yuan2011PAMI}. While action recognition is a single-label classification problem, in action detection, we can have multi-class labels for each action sequence \cite{vahdani2022PAMI, herath2017IVC}. 

\textcolor{black}{This literature review is mainly based on our main contributions, including (1) utilizing the joint representation of spatial (RGB) and motion (optical flow) modalities more effectively for action modeling; (2) using a more robust representation of the motion modality (optical flow fields); (3) designing an effective transformer, a state-of-the-art deep network, for the action modeling tasks.} 

\subsection{\textcolor{black}{Multi-modal action recognition}}
\cite{simonyan2014NIPS} is one of the first two-stream deep architectures that use both RGB images and optical flow fields and suggests that combining the optical flow and RGB images boosts the action recognition performance.  
\cite{feichtenhofer2016CVPR} improved such a two-stream network by redesigning its architecture, such as the mechanisms for feature fusion and pooling. 

Later, some researchers proposed strategies to improve the utilization of optical flow fields along with RGB images. 
For example, \cite{shi2017TMM} concluded that combining the trajectory descriptors and optical flow can enrich the action feature representation. 
\cite{ma2018PR} improved this by creating more effective fine-grained action descriptors that are located in informative small regions in RGB and optical flow frames. \cite{tu2018PR} enhanced the previous work by proposing a multi-stream convolutional neural network that extracts effective motion and spatial features centralized in the human body, which is more informative than the small regions in \cite{ma2018PR}.

Some other works have been conducted to resolve some general issues in action recognition based on two spatial and temporal modalities. 
\cite{wang2017TMM} introduced a 3-D convolutional fusion to deal with varying spatial and temporal sizes of RGB and optical flow frames. 
\cite{zhu2018ACCV} proposed a method that improves the efficiency of motion feature extraction in a two-stream convolutional network to take advantage of the joint representation of spatial and the proposed efficient motion modalities.
\cite{xiong2020JMS} suggested a pipeline to transfer the knowledge obtained from the dataset with a large volume of RGB and optical flow action frames to smaller-scale real-world scenarios such as manufacturing. 

\subsection{\textcolor{black}{Multi-modal temporal action detection}}
To enhance the way the optical flow fields and RGB frames are used for temporal action detection, some researchers suggested strategies that focus on sub-regions in images.
For example, \cite{peng2016ECCV} concluded that focusing on local regions in RGB and optical flow fields (by using a motion region network) and stacking optical flow improves the modeling of actions.  \cite{zhang2018MMTA} advanced such a spatial-only region-based approach  by focusing on semantic regions, multiple object tracking, and person detection to capture better action proposals from RGB images and optical flow frames. 
 \cite{zhang2022TVC} improved the previous work by adding a temporal dimension to the spatial one, using a temporal cuboid representation around subjects and objects in both RGB images and optical flow frames with an appearance and motion detector pre-processing step.

Some researchers proposed using optical flow fields more efficiently. 
\cite{zhang2019ICIP} suggested reducing the number of optical flow and RGB frames needed for creating effective spatial-temporal action features utilizing long-term 3D CNNs. As a better strategy, instead of compromising  the number of  frames,  \cite{chang2022PR} proposed a convolution autoencoder  to extract spatial and temporal features and effectively simulate the optical flow information by using consecutive frames.

Other works suggested jointly enhancing the optical flow field and RGB image representation.
For example, \cite{su2020PAMI} performed the tasks of spatial-temporal localization and action classification using a cross-stream cooperation strategy, where RGB and optical flow streams jointly improve these tasks. \cite{zhai2022PAMI} enhanced such as joint learning by distinguishing between the actions and background in both RGB and optical flow frames, which are weakly annotated.

\subsection{\textcolor{black}{Transformer network for action modeling}}
Some approaches suggested using a transformer network or an attention mechanism for action recognition and detection. 
For example, \cite{dai2020ASC} proposed a two-stream network using an attention module that focuses on the selection of effective spatial-temporal input features. 
\cite{kim2021PR} improved this selection by proposing a 3D CNN with an attention agent to remove the redundant temporal information.
\cite{neimark2021ICCV} upgraded such temporal attention using a video transformer network that is more effective in modeling the spatial-temporal feature representation than a 3D CNN \cite{kim2021PR}. 
\cite{mazzia2022PR} enhanced such a temporal action modeling by combining the power of attention and recurrent mechanisms to shorten the temporal window required for action recognition. \cite{dong2022PR} further improved it by including a Markov decision process to train an attention mechanism to capture key frames in action videos more effectively.
While the previous work only focused on temporal enhancement,  \cite{hu2022PAMI}  suggested a self-attention module to improve the action features in both spatial and temporal domains by capturing the interactions between different spatial-temporal feature maps.

Following the literature, our temporal action detection method also utilizes both RGB and optical flow modalities to effectively incorporate spatial and motion information. We propose a transformer network with a multi-modal attention mechanism and enhanced motion features to improve the expressive power of spatial-temporal action features using RGB images and optical flow fields.  

\section{Methodology}

\subsection{Overview and terminology}
\label{Sec:overview}
\begin{figure*}[!htbp]
	\centering 
		\includegraphics[height=0.32\textheight]{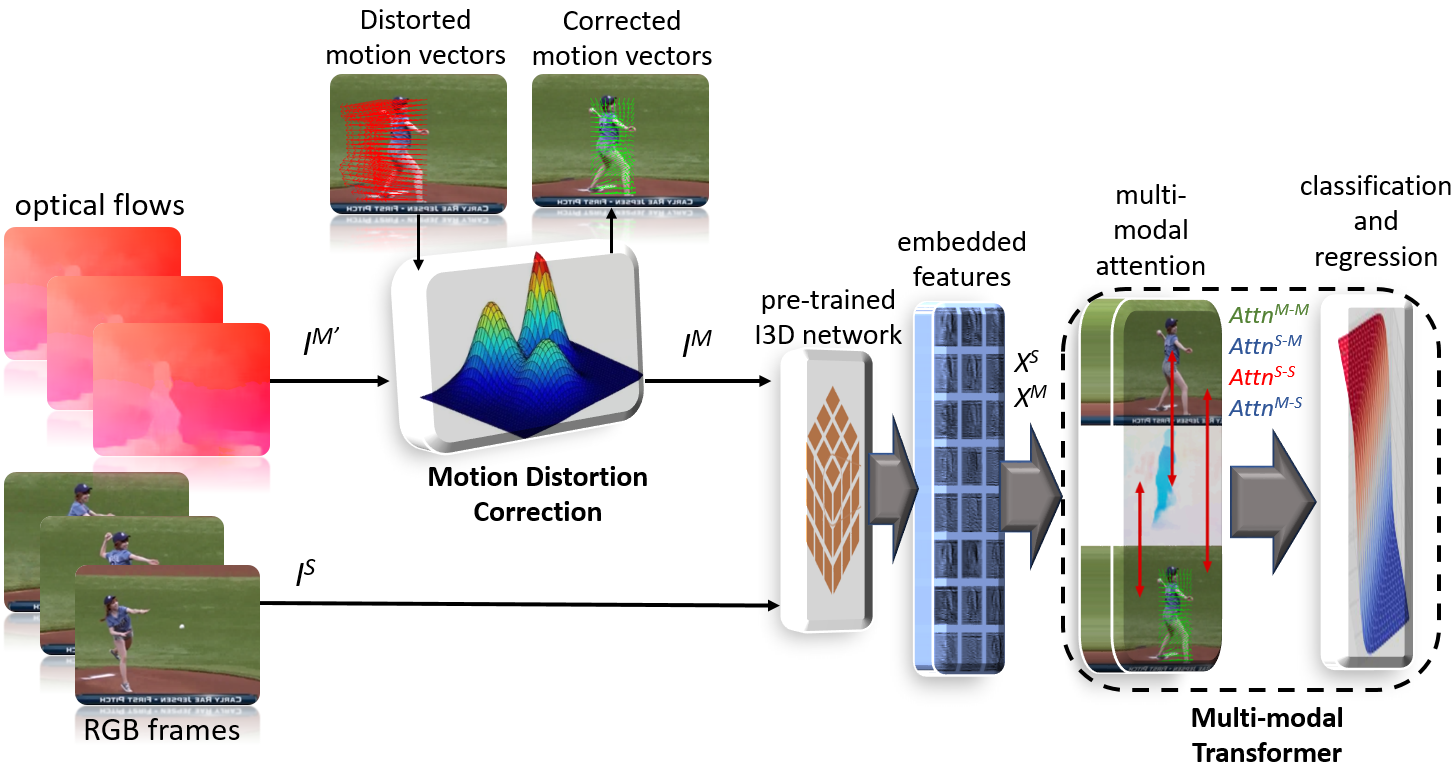} \\
	\caption{The main pipeline of our proposed action detection algorithm using both RGB frames ($I^{S}$) and optical flow fields  ($I^{M'}$). Our motion distortion correction algorithm fixes the distorted motion vectors ($I^{V'}$) and the corresponding distorted optical flow fields  ($I^{M'}$) that are caused by the camera movement and produces the corrected motion vectors ($I^V$) and the corresponding corrected optical flow images ($I^{M}$).  \textcolor{black}{The pre-trained feature embedding network, I3D \cite{carreira2017CVPR} extracts the spatial and motion features, $X^S$, and $X^M$, respectively.}  Our multi-modal transformer network uses both spatial and corrected motion modalities to detect the action classes in videos. Our multi-modal attention mechanism calculates  motion-to-motion, spatial-to-motion, motion-to-spatial, and spatial-to-spatial attention, $Attn^{M-M}$, $Attn^{S-M}$, $Attn^{M-S}$, and $Attn^{S-S}$, respectively. Finally, we identify the action sequence in the classification and regression stage. ~\label{Fig:pipe}}
\end{figure*}

Fig. \ref{Fig:pipe} indicates the  summary overview of our proposed method. Given a sequence of RGB frames, $I^S=\{I^s_i, i=0,1,...,T\}$, and corresponding distorted optical flow fields $I^{M'}=\{I^{m'}_i, i=0,1,...,T\}$, respectively, the goal is to find the action class scores, $\hat{Y}$ and the corresponding start and end of actions. Here, $T$ is the length of the temporal sequence. To do such, we first fix the distortion in the motion vectors ($I^{V'})$ corresponding to  $I^{M'}$ using our \emph{motion distortion correction} algorithm to handle camera movements. In this stage, the outputs are the corrected motion vectors, $I^V$, and corresponding corrected optical flow images, $I^{M}=\{I^{m}_i, i=0,1,...,T\}$.  
\textcolor{black}{Then, given $I^S$ and $I^M$, we embed the features using the I3D network \cite{carreira2017CVPR}, pre-trained on the  Kinetics dataset, that outputs the spatial and motion embedded features, $X^S$, and $X^M$, respectively. Following previous work  \cite{zhang2022ECCV, shi2022ECCV, liu2022TIP}, such a feature extraction (embedding) in the pre-training stage helps the network to start with an improved initialization of the action features, which is especially important for a transformer network that lacks a convolutional layer to extract features explicitly.}

Our multi-modal transformer uses $I^M$ and $I^S$ to compute the multi-modal attention types, including motion-to-motion, spatial-to-motion, motion-to-spatial, and spatial-to-spatial attention, $Attn^{M-M}$, $Attn^{S-M}$, $Attn^{M-S}$, and $Attn^{S-S}$, respectively. Then in the classification stage, we compute the action class label scores for each frame $\hat{Y}^C =\{y_i, i=0,1,...,T\}$ followed by the regression stage that calculates the start and end of actions. 
We will explain the above pipeline in detail in the following sections.

\subsection{Motion distortion correction algorithm}
\label{Sec:Dist}
In action videos captured in the wild such as those for popular untrimmed action datasets, THUMOS14 \cite{idrees2017thumos}  and ActivityNet \cite{caba2015activitynet}, that we used in our experiments, camera movements happen often. Such camera movements significantly distort the motion information depicted in optical flow fields, which is a powerful and popular modality to represent actions. We previously showed some examples of the motion distortion caused by camera movements in Fig. \ref{Fig:motion_distort}.
To solve such motion distortion and to use the optical flow fields in our multi-modal transformer effectively, we propose a novel motion distortion correction algorithm. 

Given the distorted  optical flow field images, $I^{M'}(x, y)$ and the corresponding motion vectors, $I^{V'}(u', v')$, the goal of our motion distortion correction algorithm is to effectively define a function $\psi$ so that $\psi:I^{V'}(u', v') \rightarrow I^{V}(u, v)$, where $I^{V}(u, v)$ is the set of corrected motion vectors. Here, $x$ and $y$ are image pixels, and  $u'$ and $v'$ are distorted motion displacements between the image pixels in the time $t$, as $(x^{(t)}, y^{(t)})$,  and the time $t+\tau$, as $(x^{(t+\tau)}, y^{(t+\tau)})$. $u$, and $v$ are the corrected motion displacements.

Our motion distortion correction algorithm includes three main steps: motion segmentation, background motion modeling, and motion restoration, which are explained below.

\textbf{Motion segmentation.}
In this phase, we segment the distorted motion vectors, $I^{V'}(u', v')$, to the background motion vectors, $I^{V'}_B(u', v')$, and foreground motion vectors, $I^{V'}_F(u', v')$, using a person detection algorithm \cite{carion2020ECCV}. We assume that the most important moving subjects are the persons in the scene, as actions are often defined based on persons' movements or interactions. Moreover, the persons often have dominant movements in the scene; consequently, we consider them as the foreground in action videos. 

\textbf{Background motion modeling.}
While the foreground motion is the consequence of both the camera and local movements, the often static background is mainly affected by camera movements. So, modeling the background motion is an effective way to interpret the camera movements. We use Gaussian mixture models (GMMs) for modeling motion displacement vectors of the background that may include several sub-random movements.


For the distorted background motion vectors, assuming $I^{V'}_B(s') = \{s'_n, n \in 0,..., H\}$, where $s'=(u', v')$, and $H$ is the size of $I^{V'}$, the GMMs with $M$ distributions can be formulated as:

\begin{equation}
    P(s')  = \sum_{m=1}^{M} \pi_m N(s'|\mu_m, \Sigma_m),
\end{equation}

In the above, $N(s'|\mu_m, \Sigma_m)$ is a sub-Gaussian density with the mean of $\mu_m$ and the covariance of $\Sigma_m$, weighted with the mixing coefficient of $\pi_m$. 
We model $I^{V'}_B(s')$ using the maximum likelihood estimation of the GMMs \cite{bishop2006pattern}. The algorithm is summarized as follows:

\begin{enumerate}
\item Initializing $\mu_m$, $\Sigma_m$, and $\pi_m$ 
\item Computing the posterior probability:

\begin{equation}
P(z_{nm}) = \frac{\pi_m N(s'_n|\mu_m, \Sigma_m)}{\sum_{i=1}^M \pi_i N(s'_n|\mu_m, \Sigma_m)},
\end{equation}

\item  Re-estimating the model parameters:
\begin{equation}
    \hat{\mu}_m = \frac{1}{\lambda_m} \sum_{n=1}^{H}P(z_{nm})s'_n,  \qquad \hat{\Sigma}_m = \frac{1}{\lambda_m} \sum_{n=1}^{H}P(z_{nm})(s'_n - \hat{\mu}_m)(s'_n - \hat{\mu}_m)^T, \qquad \hat{\pi_m} = \frac{\lambda_k}{H}, 
\end{equation}

where, $\lambda_m = \sum_{n=1}^{H}P(z_{nm})$

\item Obtaining the log-likelihood:

\begin{equation}
    ln P(I^{V'}_B(s')|\mu, \Sigma, \pi) = \sum_{n=1}^{H}ln(\sum_{m=1}^{M}\pi_m N(s'_n|\mu_m, \Sigma_m))
\end{equation}

\item Repeating from (2) until convergence criteria are met:  $|\hat{\mu}_m - \mu_m| \leq \theta_{\mu}$,  \quad $|\hat{\Sigma}_m - \Sigma_m| \leq \theta_{\Sigma}$,  \quad $|\hat{\pi}_m - \pi_m| \leq \theta_{\pi}$  \quad or  \quad $lnP(I^{V'}_B(s')|\mu, \Sigma, \pi) \leq \theta_{l}$

where, $\theta_\mu$, $\theta_\Sigma$, $\theta_\pi$, and $\theta_{l}$ are convergence threshold values.

\end{enumerate}

\textbf{Motion restoration.}
Previously, we modeled the camera movements using the GMMs, which we call $M^{gmm}$. To restore the motions, we assume that each motion vector $s'_n \in I^{V'}$ is affected by the camera movements' average values $\mu_m$, where $s_n$ is clustered as one of the $m \in M$ distributions as $m_n = M^{gmm}(s'_n)$. Here, we call such a clustered motion vector (to distribution $m_n$) as $s'^{(m)}_n$. The following shows how we formulate the motion restoration for each motion vector:

\begin{equation}
    for \quad s'^{(m)}_n \in I^{V'} ,  \quad s_n \in I^{V}:  \quad s_n = s'^{(m)}_n - \mu_m
\end{equation}

In the above, $s^n \in I^V$ are the corrected motion vectors. As we discussed before, our GMMs model, $M^{gmm}$, is parameterized based on 2D variables $s'_n =(u'_n,v'_n) \in  \mathbb{R}^{2}$. So, the mean variable for the distribution of $m$ is also parameterized as $\mu_m = (\mu_{u'm}, \mu_{v'm})$.

After converting the corrected motion vectors, $I^V$, to the corrected optical flow field images, $I^M$, we use the result as a more effective motion modality in the next steps of our action detection algorithm.   

\subsection{Multi-modal transformer}
\label{Sec:MMT}
\begin{figure*}[!htbp]
	\centering 
		\includegraphics[height=0.18\textheight]{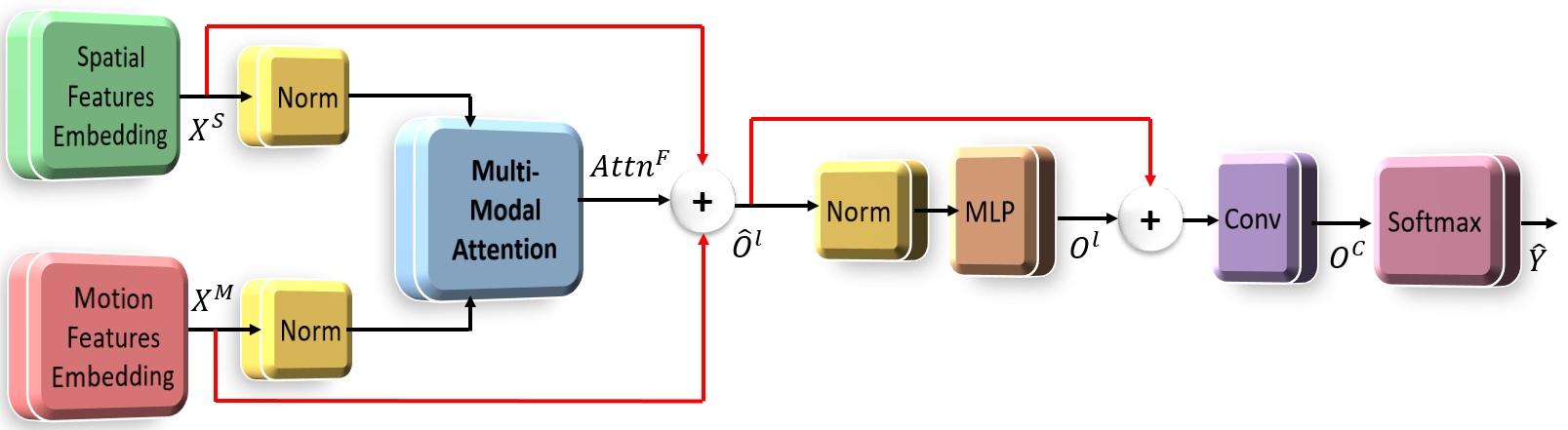} \\
	\caption{The architecture of our multi-modal transformer network includes the normalization layer (\emph{Norm}), \emph{mlti-modal attention}, \emph{MLP}, and the classification modules including the \emph{Conv}  and \emph{SoftMax} layers. \textcolor{black}{The residual connections are the red lines.} ~\label{Fig:trans}}
\end{figure*}

Fig. \ref{Fig:trans} shows the architecture of our multi-modal transformer network. The embedding $\rho$ maps the inputs $I^M$ and $I^S \in \mathbb{R}^{T\times h\times w\times 3}$ to $X^M$ and $X^S \in \mathbb{R}^{T\times Z}$, where $T$ is the temporal length, $Z$ is the embedding size, and $h$ and $w$ are frame sizes. $\rho$ is a two-stream convolutional network \cite{carreira2017CVPR} that embeds spatial and temporal (motion) modalities separately.  

We feed the embedded spatial and motion inputs, $X^M$ and $X^S$, to our multi-modal transformer network. Our transformer network includes several layers for which the relations between the consecutive layers $l-1$ and $l$ are defined as follows:

\begin{equation}
    \hat{O}^l = MMA(Norm(O^{l-1})) + O^{l-1},  \qquad l \in\{2,...,L\},
\end{equation}

\begin{equation}
    O^l = MLP(Norm(\hat{O}^{l})) + \hat{O}^{l}, \qquad l \in\{2,...,L\},
\end{equation}

In the above, $\hat{O}$ is the intermediate layer output, $O$ is the layer output, \emph{Norm} is the normalization layer, \emph{MMA} is the multi-modal attention, \emph{MLP} is a multilayer perceptron layer, and $L$ is the total number of layers. 

For the first layer, we have the following:

\begin{equation}
    \hat{O^{1}} = MMA(Norm( X^S, X^M)) + X^S + X^M,
\end{equation}

And for the final layer, we will have:

\begin{equation}
\label{EQ:class}
    \hat{Y} = p(W^F|X^S, X^M) = Softmax(Conv(O^L)),
\end{equation}

In the above, $\hat{Y}$ is the action prediction scores for each frame, $W^F$ is our network model parameters, \emph{Conv} is a convolutional layer, where $Conv:O^L \in \mathbb{R}^{T^L\times Z} \rightarrow O^C \in \mathbb{R}^C$, where $C$ is the number of classes, $O^C$ is the final output of the transformer network before the softmax layer, and $T^L$ is the temporal length of the final layer. 

\subsubsection{Multi-modal attention}
\label{Sec:MMA}
The transformer is a state-of-the-art deep network for solving spatial-temporal problems \cite{vaswani2017NIPS, neimark2021ICCV}. One of the main advantages of the transformer network is the self-attention mechanism that computes the correlative patterns among selective inputs. As we discussed before in Section \ref{Sec:intro_MMA}, finding the correlations among different spatial and motion modalities can empower the feature representation of actions.  
Hence, we propose a multi-modal attention mechanism to calculate such correlative patterns among our selective inputs, which are spatial and motion modalities, RGB images, and optical flow fields, respectively. Specifically, our multi-modal attention accommodates the modeling of a variety of actions with static and moving subjects in the scene effectively by proposing several variations in attention. Some examples are shown in Fig. \ref{Fig:att_examples}.
The first scenario is when both subjects/objects of interest (SOI) move. An example of such an action ``dancing'' is shown in Fig. \ref{Fig:att_examples} - (a), where two persons move toward each other. We compute such correlations using our motion-to-motion attention ($Attn^{M-M}$)  within our transformer network using our novel multi-modal attention mechanism. 
In the next scenario, our spatial-to-spatial attention ($Attn^{S-S}$) computes the correlative patterns in the scene when all the SOIs are static (stationary), such as the persons and table in the action ``sitting at a table'' shown in Fig. \ref{Fig:att_examples} - (b).
Next is when we have both moving and stationary SOIs in the scene that jointly define the action, such as ``kicking a ball'' (Fig. \ref{Fig:att_examples} - (c)). Here, our motion-to-spatial attention ($Attn^{M-S}$) calculates the correlations between the moving ball and the static person. And finally, our  spatial-to-motion attention ($Attn^{S-M}$)  finds the correlative patterns between the static person and moving hand (while holding glasses) in the action ``offering drink'' in Fig. \ref{Fig:att_examples} - (d).

\begin{figure*}[!htbp]
	\centering 
		\includegraphics[height=0.35\textheight]{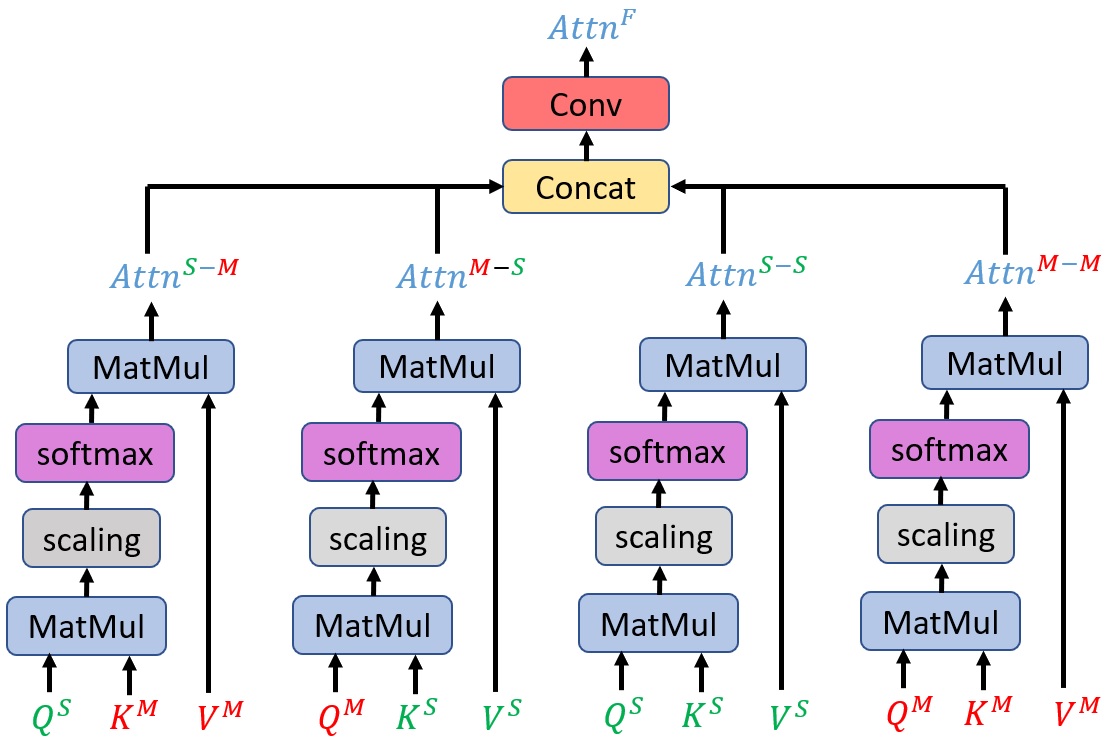} \\
	\caption{Our multi-modal attention mechanism computes the correlative patterns among spatial (RGB images) and motion (optical flow) modalities. It includes various multi-modal attention types such as  motion-to-motion ($Attn^{M-M}$), spatial-to-motion ($Attn^{S-M}$), motion-to-spatial ($Attn^{M-S}$), and spatial-to-spatial ($Attn^{S-S}$) to model a variety of actions with stationery and moving subjects effectively. ~\label{Fig:MMA}}
\end{figure*}

The structure of our multi-modal attention is illustrated in Fig. \ref{Fig:MMA}. The attention mechanism is defined as finding the correlations between the selective input, \emph{query} ($Q$), and other input candidates, \emph{keys} ($K$), which then gives us the mapped correlative results, \emph{values} ($V$). To find the correlative patterns among each modality, we introduce four attention classes: (1) spatial-to-spatial attention, $Attn^{S-S}$ computes the correlations between the spatial query, $Q^S$,  and spatial keys, $K^S$, mapped to the spatial values, $V^S$;
(2) The motion-to-motion attention ($Attn^{M-M}$) is obtained by mapping the correlations between motion query ($Q^M$) and motion keys ($K^M$) to motion values ($V^M$); 
(3) The spatial-to-motion attention ($Attn^{S-M}$) is calculated by first finding the correlations between $Q^S$ and $K^M$, which then is mapped to $V^M$; (4) The motion-to-spatial attention ($Attn^{M-S}$) is obtained similarly, but the query, keys, and values for the two modalities are switched. 

We formulate the query, keys, and values for both modalities as follows:

\begin{equation}
    Q^S = X^SW_q^S, \qquad K^S = X^SW_k^S, \qquad  V^S = X^SW_v^S,
\end{equation}

\begin{equation}
    Q^M = X^MW_q^M, \qquad K^M = X^MW_k^M, \qquad  V^M = X^MW_v^M,
\end{equation}

In the above $W_q^S \in \mathbb{R}^{Z\times Z_q}$ is the spatial query projection weights,  $W_k^S \in \mathbb{R}^{Z\times Z_k}$ is the spatial keys projection weights and $W_v^S \in \mathbb{R}^{Z\times Z_v}$ is the spatial values projection weights, and $Z_q$, $Z_k$, and $Z_v$ are the projection sizes for query, keys, and values, respectively. 
$W_q^M \in \mathbb{R}^{Z\times Z_q}$ indicates the motion query projection weights, $W_k^M \in \mathbb{R}^{Z\times Z_k}$ is the motion keys projection weights, and $W_v^M\in \mathbb{R}^{Z\times Z_v}$ is the motion values projection weights. 

The four multi-modal attention types (following  Fig. \ref{Fig:MMA}) are formulated as follows:

\begin{equation}
    Attn^{S-S} = Softmax(\frac{Q^S(K^S)^T}{\sqrt{Z_m}})V^S,
\end{equation}

\begin{equation}
    Attn^{S-M} = Softmax(\frac{Q^S(K^M)^T}{\sqrt{Z_m}})\textcolor{black}{V^M},
\end{equation}

\begin{equation}
    Attn^{M-S} = Softmax(\frac{Q^M(K^S)^T}{\sqrt{Z_m}})\textcolor{black}{V^S},
\end{equation}

\begin{equation}
    Attn^{M-M} = Softmax(\frac{Q^M(K^M)^T}{\sqrt{Z_m}})V^M,
\end{equation}

In the above, $Z_m$ is the model size. The details for all the aforementioned parameter values are explained in Section \ref{Sec:Impl}.

The final attention, $Attn^{F}$ is obtained as:
\begin{equation}
    Attn^{F} = Conv(Concat(Attn^{S-S}, Attn^{S-M}, Attn^{M-S}, Attn^{M-M})),
\end{equation}

In the above $Conv$ is a convolutional layer where $Conv: \mathbb{R}^{4 \times T\times Z} \rightarrow \mathbb{R}^{T\times Z}$ and $Concat$ is a concatenation operator. 

\textcolor{black}{\cite{song2023arXiv} suggested an approach for correlative masked modeling. The correlations are computed between the search and template sub-images for tracking purposes. In their proposed method so-called packed self-attention (PSelf-Attn) the search and template tokens are concatenated  before computing the attention maps. Their attention mechanism  will provide the cross-attention maps that we suggested in our paper. The number of operations still remains similar to our method.}

Our network's loss function is shown as follows:

\begin{equation}
    L = - \sum_{t=1}^{T} \sum_{c = 1}^{C} y_t^{(c)} log \hat{y}_t^{(c)} + \alpha Loss_{tIOU}
\end{equation}

In the above, $y$ and $\hat{y}$ are ground truth and predicted values for each video frame and class $c$ and time $t$, respectively. $Loss_{tIOU}$ is the temporal intersection loss that indicates the similarities between the predicted video segment frames and positive ground truth frames for the duration of $1 \leq t \leq T$. $\alpha$ is a loss adjustment term.

\subsubsection{Classification and regression}
\textcolor{black}{Given the action detection scores, $\hat{Y}$, calculated following Equation \ref{EQ:class}, the goal of the regression stage is to find the start and end frames of action instances, $\{\Delta, E\} = \{\delta_{n, c}, \epsilon_{n, c},  n = 0,1...,N'; c = 0,1,...C\}$, where $N'$ is the number of segmented class instances. Moreover, $\delta_{n, c}$ and $\epsilon_{n, c}$ are the start and end frames for class $c$ of instance $n$. Our regression algorithm consists of two phases of action scores thresholding and the start and end frames detection as shown in Algorithm \ref{alg:se}. }

\begin{algorithm}
\begin{algocolor}
\setstretch{1.0}
\caption{Regression}\label{alg:se}
\begin{algorithmic}[1]
\Require Action detection scores, $\hat{Y}\in \mathbb{R}^{T \times C}$ 
\Ensure start and end frames of action instances, $\Delta, E$ 
\phase{action scores thresholding}
\State $t = 0 $
\While{$t \leq T$} \Comment{$T$ is \# of frames}
\State $c = 0 $
\While{$c \leq C$} \Comment{$C$ is \# of classes}
\If{$\hat{Y}_{t,c} \geq \theta$} \Comment{$\theta$ is the detection threshold}
\State $\hat{Y}_{t,c} = 1$
\ElsIf{$\hat{Y}_{t,c} < \theta$}
\State $\hat{Y}_{t,c} = 0$
\EndIf
\State $c \gets c+1$  
\EndWhile
\State $t \gets t+1$  
\EndWhile
\phase{Start and end frames detection}
\State $Y^S = SPLIT(\hat{Y})$ \Comment{splitting $\hat{Y}$ to $N$ segments}
\State $Y^S = \{y_n, n = 1,2,..., N\}$
\State $L \in \mathbb{R}^{N \times C} \gets 0$ \Comment{initializing labels for each segment} 
\State $n = 0 $
\While{$n \leq N$} 
\State $c = 0 $
\While{$c \leq C$} 
\If{$\sum_{i=1}^{i=Q} y_{n,i,c} \geq Q/2$} \Comment{$Q = T/N$}
\State $L_{n,c} = 1$
\ElsIf{$\sum_{i=1}^{i=Q} y_{n,i,c} < Q/2$}
\State $L_{n,c} = 0$
\EndIf
\State $n \gets n+1$  
\EndWhile
\State $c \gets c+1$ 
\EndWhile
\State $L^M = MERGE(L)$ \Comment{merging if $L_{n,c} = L_{n+1,c}$}
\State $Y^M = RESPLIT(\hat{Y})$ \Comment{re-splitting $\hat{Y}$ based on $L^M$} 
\State $n = 0 $
\While{$n \leq N'$} \Comment{$N'$ is \# of segments for $L^M$}
\State $c = 0 $
\While{$c \leq C$} 
\State   $\delta_{n,i,c} = START(Y^M_{n,i,c})$ \Comment{$i$: 1st time $Y^M_{n,i,c} = 1$}
\State   $\epsilon_{n,j,c} = END(Y^M_{n,j,c})$ \Comment{$j$: last time $Y^M_{n,j,c} = 1$}
\State Add $\delta_{n,i,c}$ to $\Delta$ \Comment{adding the start frame}
\State Add $\epsilon_{n,j,c}$ to $E$ \Comment{adding the end frame}
\State $n \gets n+1$  
\EndWhile
\State $c \gets c+1$ 
\EndWhile
\end{algorithmic}
\end{algocolor}
\end{algorithm}

\subsection{Implementation details}
\label{Sec:Impl}

Table \ref{Tab:impl} indicates the implementation details of our proposed pipeline. All the experiments are conducted using PyTorch 1.7 on a server PC with dual Nvidia RTX 3090 GPUs (24GB VRAM), AMD Ryzen Threadripper 3990X 64-Core Processor, and 256GB of RAM.

\begin{table}[h!tbp]
	\centering
	\caption{Implementation details of our proposed pipeline with associated paper section references.} \label{Tab:impl} 
\begin{tabular}{ccc}
\hline
Parameter & Value & Section \\ \hline
Number of GMM distributions ($M$) & 16 & \ref{Sec:Dist} \\
Spatial and Temporal Features Embedding Size ($Z$)  & 1024 & \ref{Sec:MMT} \\ 
Number of Transformer Layers ($L$) & 6 &  \ref{Sec:MMT} \\
Kernel Size for \emph{Conv} & 3 & \ref{Sec:MMT} \\
Learning Rate & $1e^{-5}$ &  \ref{Sec:MMT} \\
Number of Training Epochs & 100 & \ref{Sec:MMT} \\
Optimizer & ADAM & \ref{Sec:MMT} \\
Weight Decay & $1e^{-6}$ & \ref{Sec:MMT}\\
Maximum Temporal Window Length ($T$) & 2304 & \ref{Sec:MMT} \\ 
Projection Size for Query and Keys ($Z_q$, and $Z_k$) & 512 & \ref{Sec:MMA}\\ 
Projection Size for Values ($Z_v$) & 1024 & \ref{Sec:MMA}\\ 
Number of MMA Heads & 3 & \ref{Sec:MMA} \\ 
Model size ($Z_m$) & 512 & \ref{Sec:MMA} \\ 
Loss Adjustment Term ($\alpha$) & 1  & \ref{Sec:MMA} \\ 
\hline
\end{tabular}
\end{table}

\section{Experimental Results}

\subsection{Datasets and experimental setup}

\subsubsection{Public benchmarks}
We used three THUMOS14 \cite{idrees2017thumos}, ActivityNet \cite{caba2015activitynet}, and our collected instructional activity datasets in our experiments. THUMOS14 and ActivityNet datasets are among the most well-known untrimmed activity datasets that have been used widely for action detection \cite{lin2021CVPR, zhao2021ICCV, zeng2019ICCV}. THUMOS14 consists of 413 untrimmed videos of 20 action classes. Following \cite{lin2021CVPR, zhao2021ICCV, zeng2019ICCV}, we used 200 videos for training and 213 videos for testing. 
ActivityNet consists of 20,000 videos of 200 action classes. Following \cite{lin2021CVPR, zhao2021ICCV, zeng2019ICCV}, we used 10,024 videos for training and 4,926 videos for testing. 

\subsubsection{Instructional activity dataset}
\label{Sec:classroom_dataset}
\begin{figure*}[!htbp]
	\centering 
		\includegraphics[height=0.23\textheight]{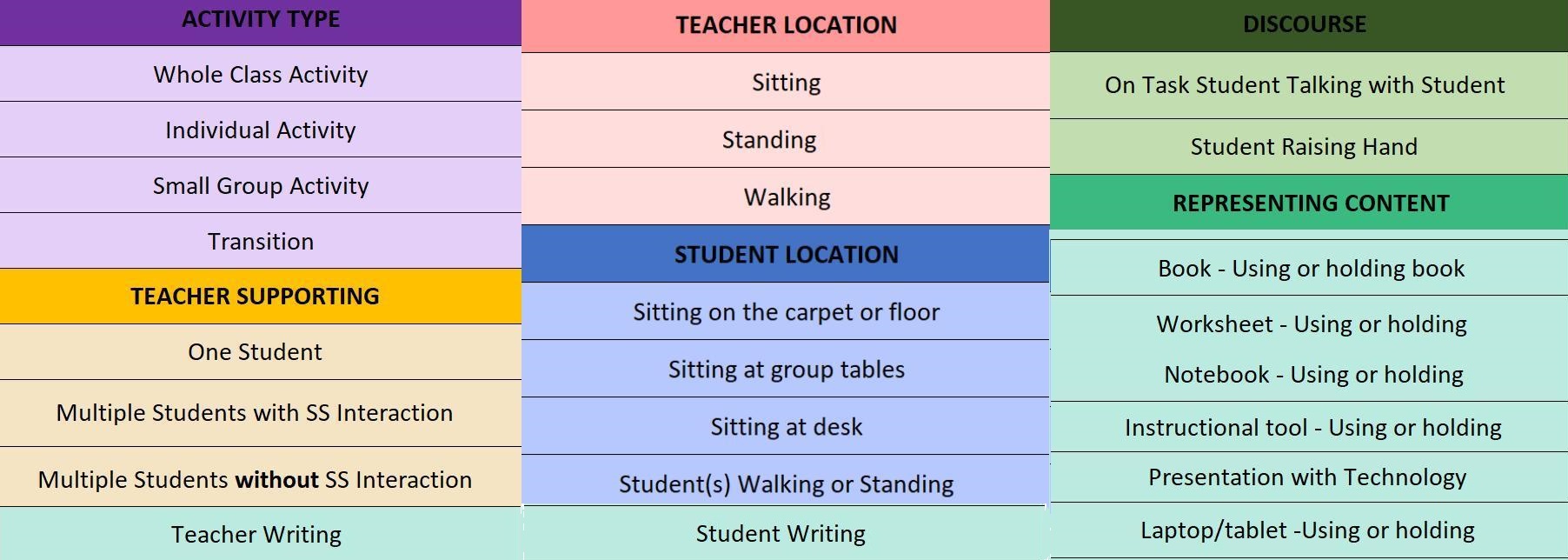} \\
	\caption{Acivity class labels of our instructional activity dataset. ~\label{Fig:our_labels}}
\end{figure*}

\begin{figure*}[!htbp]
\renewcommand{\tabcolsep}{0.5pt}
	\centering 
		\begin{tabular}{ccc}
		\includegraphics[height=0.12\textheight]{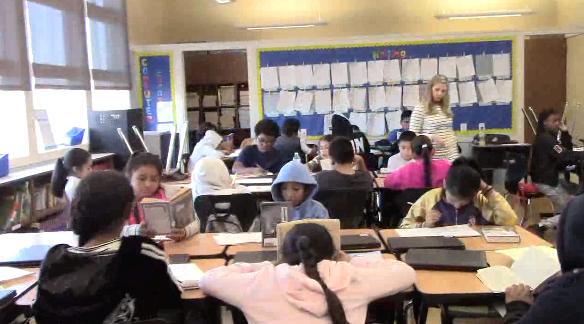}  & 
            \includegraphics[height=0.12\textheight]{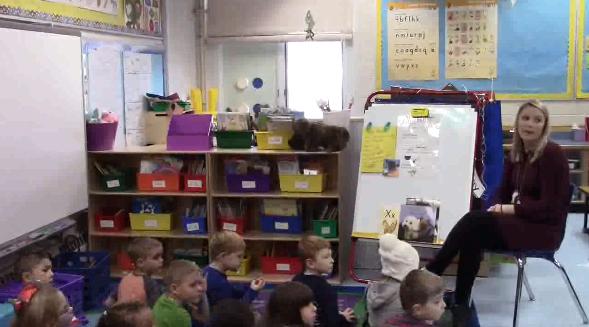} &
            \includegraphics[height=0.12\textheight]{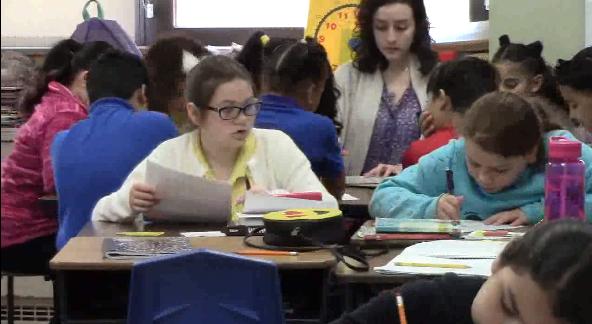} \\
            \includegraphics[height=0.12\textheight]{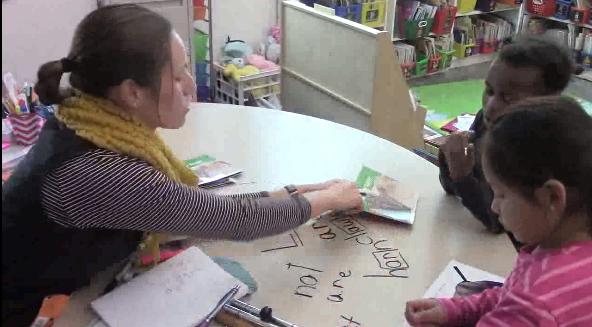} &
            \includegraphics[height=0.12\textheight]{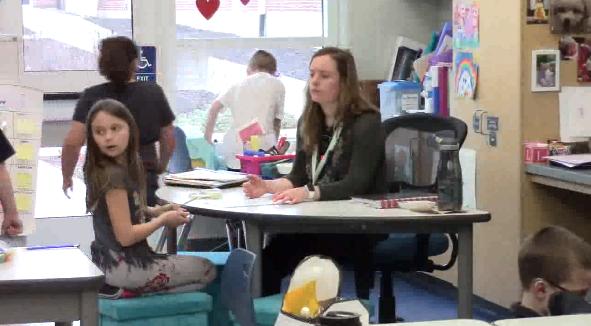} &
            \includegraphics[height=0.12\textheight]{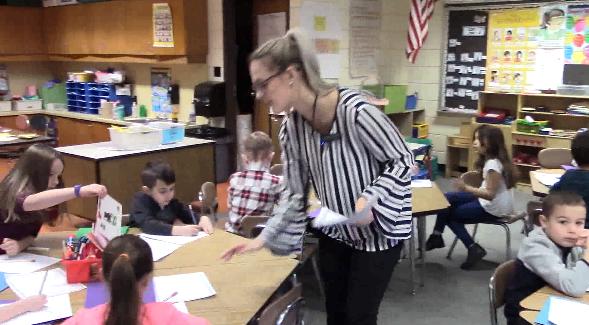} \\

    	\end{tabular}
	\caption{Some example frames of our instructional activity dataset.  ~\label{Fig:class_exms}}
\end{figure*}

\begin{figure*}[!htbp]
\renewcommand{\tabcolsep}{0.5pt}
	\centering 
		\begin{tabular}{ccc}
		\includegraphics[height=0.12\textheight]{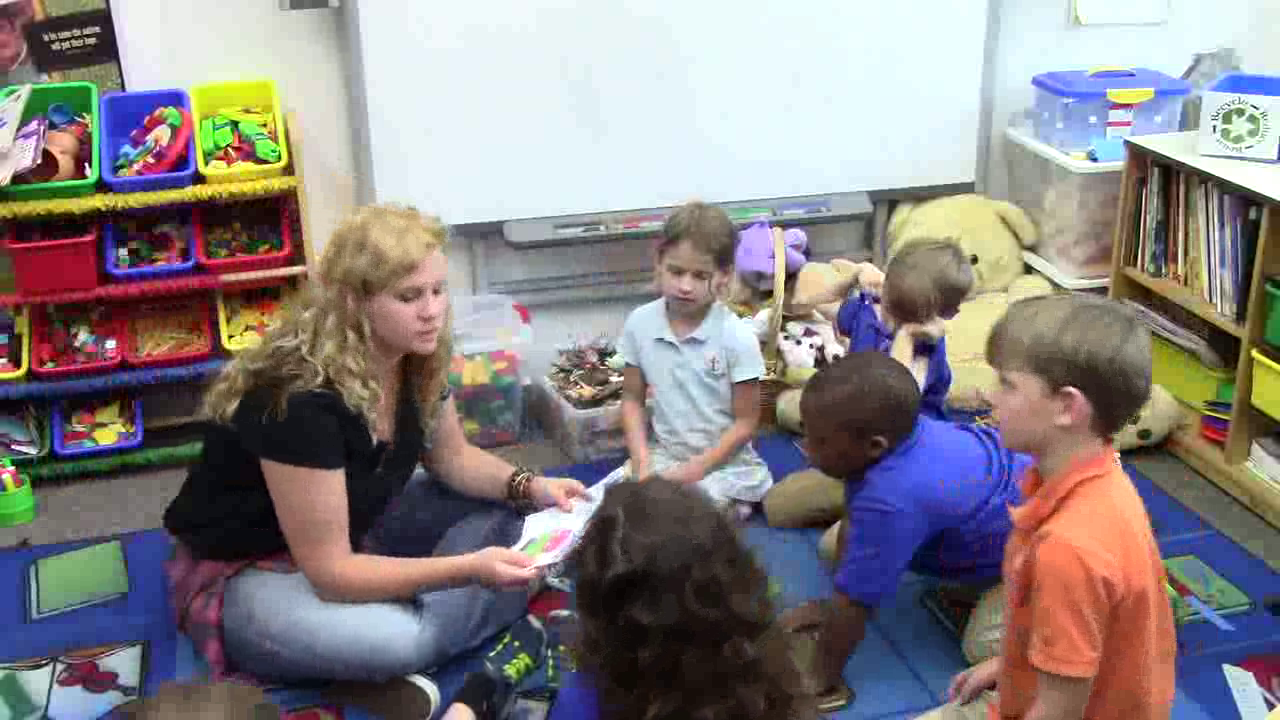} &
  \includegraphics[height=0.12\textheight]{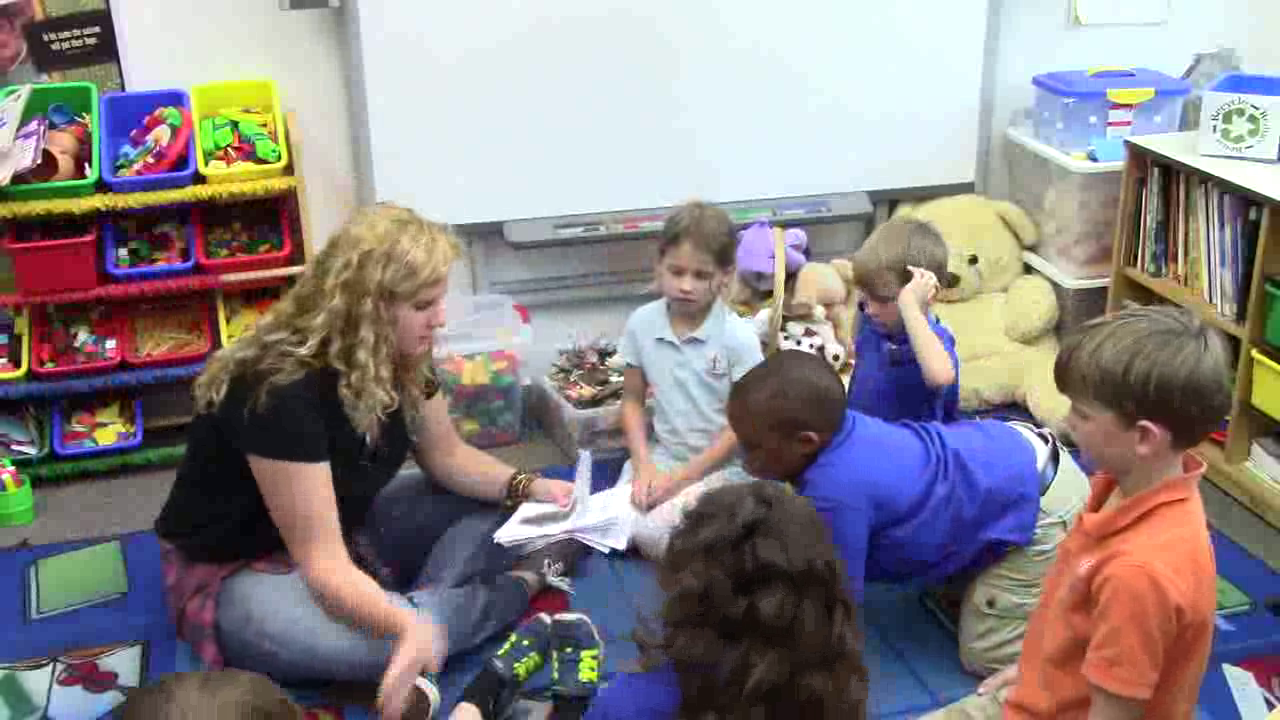} &
  \includegraphics[height=0.12\textheight]{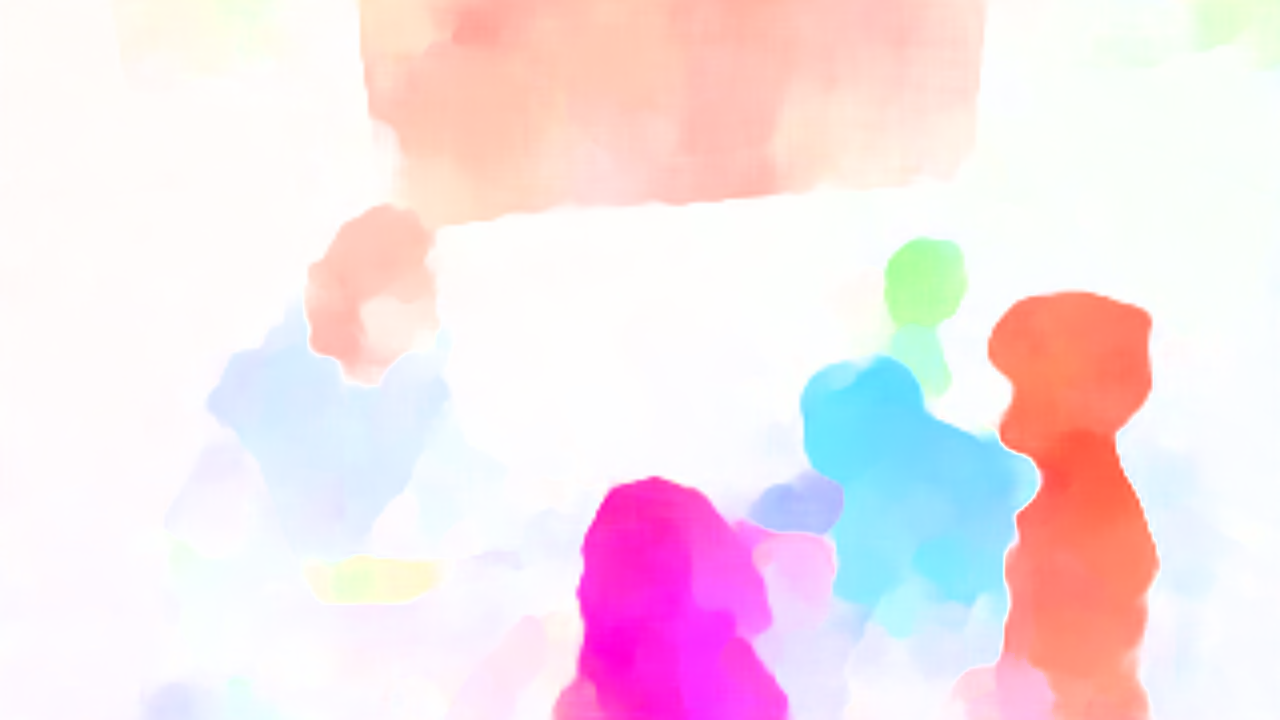} \\
  \includegraphics[height=0.12\textheight]{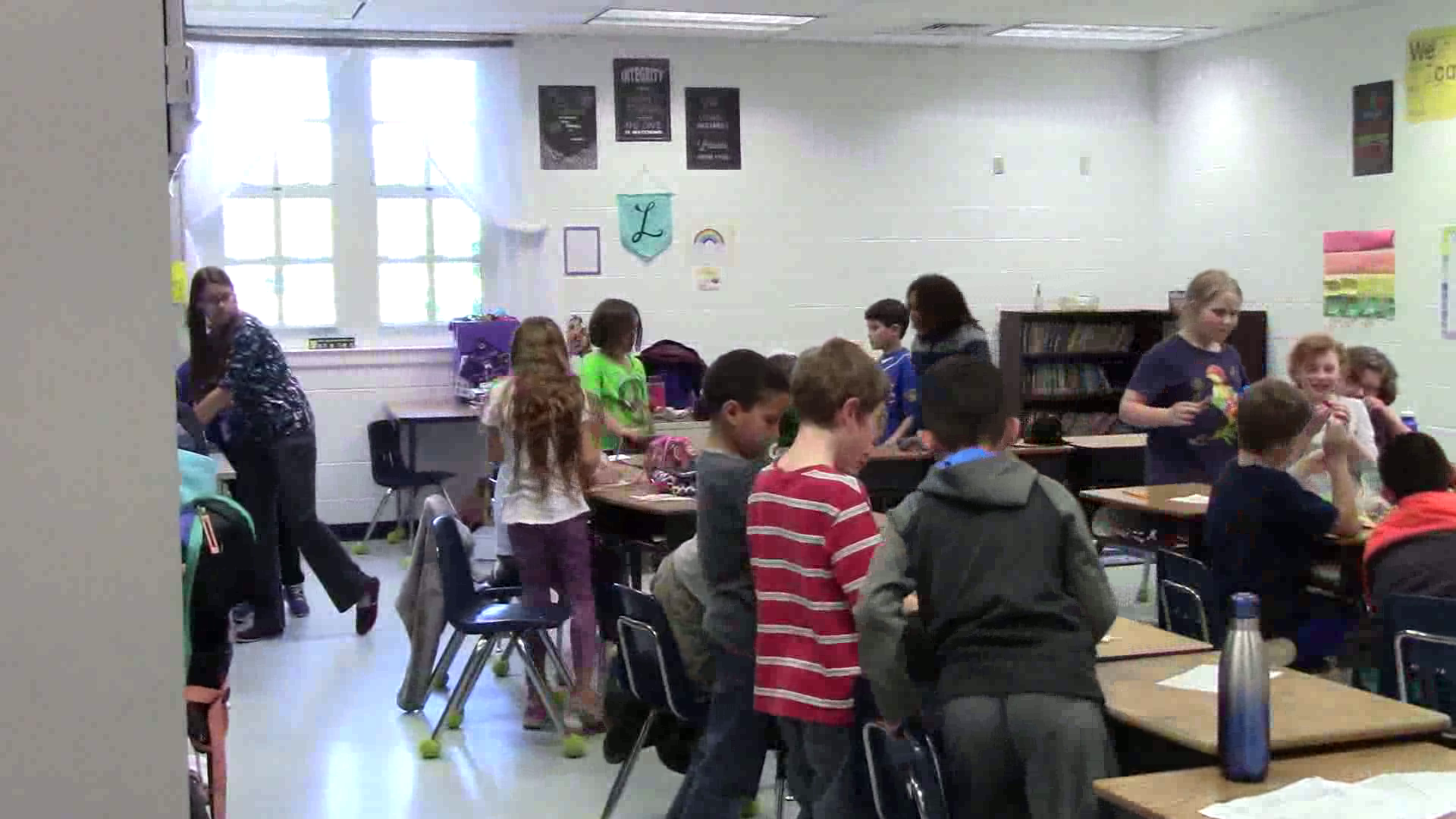} &
  \includegraphics[height=0.12\textheight]{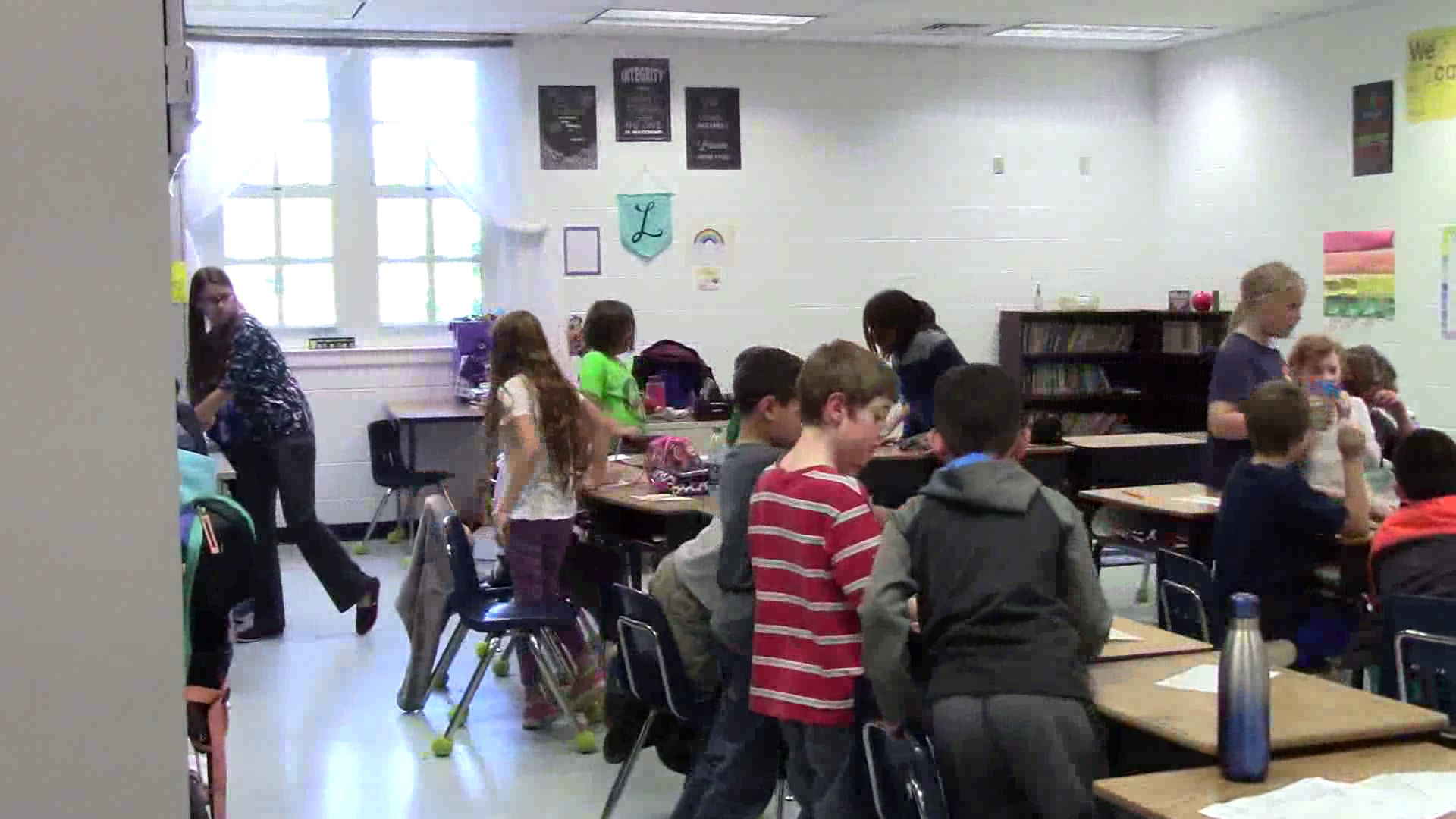} &
  \includegraphics[height=0.12\textheight]{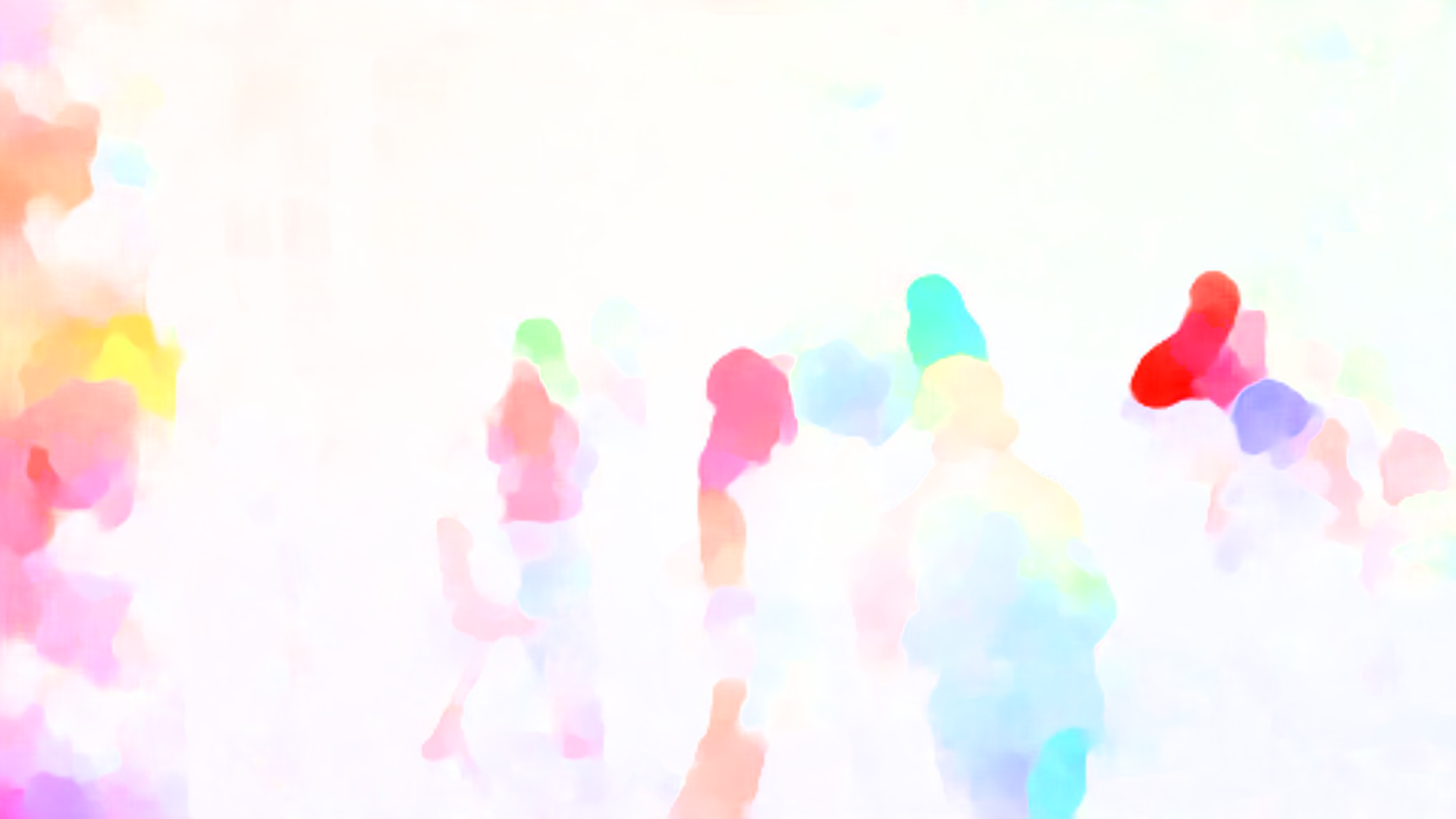} \\
  frame $t$ & frame $t + \tau$ & optical flow \\ 

    	\end{tabular}
	\caption{Some example frames of our annotated instructional activity dataset showing two consecutive sampled frames of $t$ and frame $t + \tau$, and consequently, the computed optical flows. ~\label{Fig:class_dataset}}
\end{figure*}

We created a dataset of instructional activities recorded from elementary schools. We annotated 240 hours of instructional activity videos with a professional team of nine annotators. Our 24 instructional activity class labels are shown in Fig. \ref{Fig:our_labels}. Some frame examples of our instructional activity dataset are shown in Fig. \ref{Fig:class_exms}.
We annotate every second (30 frames) of the video  with multiple class labels. Some frame examples from our annotated videos and computed optical flow are shown in Fig. \ref{Fig:class_dataset}. 
The public link to download our dataset will be on our website \cite{AIAI} when it is available online. 
In this experiment, we used 50 hours of our videos with training and testing set proportions of 80\% and 20\%, respectively. 

\subsubsection{Evaluation Metric} We used the mean average precision (mAP) at different thresholds of temporal intersection over union (tIoU), which is the most common metric used in action detection. For the THUMOS14 and ActivityNet, we reported the results for the threshold sets of \{0.3, 0.4, 0.5, 0.6, 0.7\} and \{0.5, 0.75, 0.95\}, respectively. The aforementioned thresholds are the standard benchmarks that have been used for these two datasets in the literature \cite{lin2021CVPR, zhao2021ICCV, zeng2019ICCV}.

\subsection{Comparative results on public datasets}
We compared our methods with the state-of-the-art strategies including  \textbf{AF (ECCV 2022)} \cite{zhang2022ECCV}, \textbf{ReAct (ECCV 2022)}, \cite{shi2022ECCV}, \textbf{TadTR (TIP 2022)} \cite{liu2022TIP}, \textbf{AFSD (CVPR 2021)} \cite{lin2021CVPR}, \textbf{VSGN (ICCV 2021)} \cite{lin2021CVPR}, \textbf{BMN-CSA (ICCV 2021)} \cite{sridhar2021ICCV}, \textbf{TCANet (CVPR 2021)} \cite{qing2021CVPR}, \textbf{MUSES (CVPR 2021)} \cite{liu2021CVPR}, \textbf{TSA-Net (CVPR 2021)} \cite{qing2021CVPR}, \textbf{ RTD-Net (ICCV 2021)} \cite{tan2021ICCV} , \textbf{TAL-MR (ECCV 2020)} \cite{zhao2020ECCV}, \textbf{A2Net (TIP 2020)} \cite{yang2020TIP}, \textbf{BMN (ICCV 2019)}, \cite{lin2019ICCV}, and \textbf{P-GCN (ICCV 2019)} \cite{zeng2019ICCV}. The comparative results for the THUMOS14 and ActivityNet datasets are shown in Table \ref{Tab:compare_thu} and Table \ref{Tab:compare_act}, respectively. As can be seen, our method outperformed the state-of-the-art approaches on these two public benchmarks based on different mAP thresholds. 
Specifically, for the THUMOS14 dataset, our proposed method outperformed the current best benchmark, AF \cite{zhang2022ECCV}, by 3.1\%, 2.2\%, 2.4\%, 2.3\%, and 1.4\% at mAP of 0.3, 0.4, and 0.7, respectively. For the ActivityNet dataset, our method suppressed the best one AF \cite{zhang2022ECCV} by  3.4\%, 1.7\%, and 0.7\% at the mAP of 0.5, 0.75, and 0.95, respectively.

\begin{table}[h!tbp]
	\centering
	\caption{Comparison of our method, and the state-of-the-art methods on the THUMOS14 dataset.} \label{Tab:compare_thu} 
\resizebox{\textwidth}{!}{\begin{tabular}{cccccccc}
\hline
Team (Year) & Method & maP@0.3 & maP@0.4 & maP@0.5 & maP@0.6 & maP@0.7 & Avg \\ \hline
\cite{yang2020TIP} (2020) & A2Net & 58.6 & 54.1 & 45.5 & 32.5 & 17.2 & 41.6 \\ \hline
\cite{zhao2020ECCV} (2020) & TAL-MR & 53.9 & 50.7 & 45.4 & 38.0 & 28.5 & 43.3 \\ \hline
\cite{tan2021ICCV} (2021) & RTD-Net & 68.3 & 62.3 & 51.9 & 38.8 & 23.7 & 49.0 \\ \hline
\cite{zeng2019ICCV} (2019) & P-GCN &  69.1 & 63.3 & 53.5 & 40.4 & 26.0 & 50.5 \\ \hline
\cite{qing2021CVPR} (2021) & TSA-Net & 60.6 & 53.2 & 44.6 & 36.8 & 26.7 & 44.3 \\ \hline
\cite{liu2022TIP} (2022) & TadTR & 62.4 & 57.4 & 49.2 & 37.8 & 26.3 & 46.6 \\ \hline
\cite{liu2021CVPR} (2021) & MUSES & 68.9 & 64.0 & 56.9 & 46.3 & 31.0 & —  \\ \hline
\cite{lin2019ICCV} (2019) & BMN & 56.0 & 47.4 & 38.8 & 29.7 & 20.5 & 38.5 \\ \hline
\cite{qing2021CVPR} (2021) & TCANet & 60.6 & 53.2 & 44.6 & 36.8 & 26.7 & 44.3 \\ \hline
\cite{sridhar2021ICCV} (2021) & BMN-CSA & 64.4 & 58.0 & 49.2 & 38.2 & 27.8 & 47.7 \\ \hline
\cite{zhao2021ICCV} (2021) & VSGN & 66.7 & 60.4 & 52.4 & 41.0 & 30.4 & 50.2 \\ \hline
\cite{lin2021CVPR} (2021) & AFSD & 67.3 & 62.4 & 55.5 & 43.7 & 31.1 & 52.0  \\ \hline
\cite{shi2022ECCV} (2022) & ReAct & 69.2 & 65.0 & 57.1 & 47.8 & 35.6 & 55.0 \\ \hline
\cite{zhang2022ECCV} (2022) & AF & 82.1 & 77.8 & 71.0& 59.4 & 43.9 & 66.8 \\ \hline
\textbf{Us} & \textbf{Ours (MMNet)} & \textbf{85.2} & \textbf{80.0} & \textbf{73.4} & \textbf{61.7} & \textbf{45.3} & \textbf{68.5} \\ \hline
\end{tabular}}
\end{table}

\begin{table}[h!tbp]
	\centering
	\caption{Comparison of our method, and the state-of-the-art methods on the ActivityNet dataset.} \label{Tab:compare_act} 
\begin{tabular}{cccccc}
\hline
Team (Year) & Method & maP@0.5 & maP@0.75 & maP@0.95 & Avg \\ \hline
\cite{yang2020TIP} (2020) & A2Net & 43.6 & 28.7 & 3.7 & 27.8 \\ \hline
\cite{zhao2020ECCV} (2020) & TAL-MR & 43.5 & 33.9 & 9.2 & 30.2 \\ \hline
\cite{tan2021ICCV} (2021) & RTD-Net & 47.2 & 30.7 & 8.6 & 30.8 \\ \hline
\cite{zeng2019ICCV} (2019) & P-GCN & 48.3 & 33.2 & 3.3 & 31.1 \\ \hline
\cite{qing2021CVPR} (2021) & TSA-Net & 48.7 & 32.0 & 9.0 & 31.9 \\ \hline
\cite{liu2022TIP} (2022) & TadTR & 49.1 & 32.6 & 8.5 & 32.3 \\ \hline
\cite{liu2021CVPR} (2021) & MUSES & 50.0 & 35.0 & 6.6 & 34.0  \\ \hline
\cite{lin2019ICCV} (2019) & BMN & 50.1 & 34.8 & 8.3 & 33.9 \\ \hline
\cite{qing2021CVPR} (2021) & TCANet & 52.3 & 36.7 & 6.9 & 35.5 \\ \hline
\cite{lin2021CVPR} (2021) & AFSD & 52.4 & 35.3 & 6.5 & 34.4 \\ \hline
\cite{zhao2021ICCV} (2021) & VSGN & 52.4 & 36.0 & 8.4 & 35.1 \\ \hline
\cite{sridhar2021ICCV} (2021)  & BMN-CSA & 52.4 & 36.2 & 5.2 & 35.4 \\ \hline
\cite{shi2022ECCV} (2022) & ReAct & 49.6 & 33.0 & 8.6 & 32.6 \\ \hline
\cite{zhang2022ECCV} (2022) & AF & 54.7 & 37.8 & 8.4 & 36.6 \\ \hline
\textbf{Us} & \textbf{Ours (MMNet)} & \textbf{58.1} & \textbf{39.5} & \textbf{9.1} & \textbf{39.0} \\ \hline
\end{tabular}
\end{table}

\subsection{Ablation study}
We conducted an ablation study to evaluate the impact of the constituent components of our proposed method on the overall action detection performance on the THUMOS14 and AcivityNet datasets.


\textcolor{black}{Table \ref{Tab:attn_all} shows the impact of various multi-modal attention types on the overall action detection performance on the THUMOS14 dataset.}
\textcolor{black}{As can be seen among the intra-modality attention types, the spatial-to-spatial attention, $Attn^{S-S}$, slightly led to a better performance than the motion-to-motion attention, $Attn^{M-M}$ with an average mAP difference of 0.3\%. On the other hand, the cross-modality attention types, spatial-to-motion $Attn^{S-M}$, and motion-to-spatial $Attn^{M-S}$, resulted in competitive performance compared to other types of attention types with an average mAP of 65.9\% and 65.4\%, respectively, Using all types of attention jointly, however, led to the maximum overall action detection performance with an average mAP of 68.6\%.
Table 5 indicates the effect of different multi-modal attention mechanisms based on the ActivityNet dataset. Similarly, $Attn^{S-S}$, marginally was better than $Attn^{M-M}$ with an average mAP of 0.5\%. Among the cross-modality attention types $Attn^{S-M}$ and $Attn^{M-S}$, led to a fair average mAP of 34.1\% and 34.0\%, respectively. The maximum overall performance is achieved when all the attention types are used with an average mAP of 39.0\%. }

\begin{table}[h!tbp]
	\centering
	\caption{Impact of different types of attention types and their combinations on the overall action detection performance on the THUMOS14 dataset. These include motion-to-motion $Attn^{M-M}$, spatial-to-motion $Attn^{S-M}$, motion-to-spatial $Attn^{M-S}$, and spatial-to-spatial $Attn^{S-S}$} \label{Tab:attn_all} 
\resizebox{\textwidth}{!}{\begin{tabular}{ccccccc}
\hline
Attention type & maP@0.3 & maP@0.4 & maP@0.5 & maP@0.6 & maP@0.7 & Avg \\ \hline 
$Attn^{S-S}$ & 80.6 & 76.5 & 70.6 & 58.3 & 41.8 & 64.7 \\ \hline
$Attn^{M-M}$ & 80.3 & 76.2 & 70.3 & 58.0 & 41.5 & 64.4 \\ \hline
$Attn^{S-M}$ & 83.4 & 78.3 & 71.7 & 60.5 & 43.8 & 65.9 \\ \hline
$Attn^{M-S}$ & 83.3 & 78.4 & 71.5 & 60.2 & 43.3 & 65.4 \\ \hline
$Attn^{S-S}$ + $Attn^{M-M}$  & 83.6 & 78.7 & 72.1 & 60.9 & 44.2 & 66.4 \\ \hline 
$Attn^{S-S}$ + $Attn^{M-M}$ +  & \textbf{85.2} & \textbf{80.0} & \textbf{73.4} & \textbf{61.7} & \textbf{45.3} & \textbf{68.5} \\
$Attn^{S-M}$ + $Attn^{M-S}$ & & & & &  \\ \hline 
\end{tabular}}
\end{table}

\begin{table}[h!tbp]
	\centering
	\caption{\textcolor{black}{Impact of different types of attention types and their combinations on the overall action detection performance on the ActivityNet dataset. These include motion-to-motion $Attn^{M-M}$, spatial-to-motion $Attn^{S-M}$, motion-to-spatial $Attn^{M-S}$, and spatial-to-spatial $Attn^{S-S}$}}  \label{Tab:attn_all_act} 
\begin{tabular}{ >{\color{black}}c >{\color{black}}c >{\color{black}}c >{\color{black}}c >{\color{black}}c}
\hline
Attention type & maP@0.5 & maP@0.75 & maP@0.95 & Avg \\ \hline 
$Attn^{S-S}$ & 52.8 & 33.2 & 7.1 & 32.9 \\ \hline
$Attn^{M-M}$ & 51.1 & 33.1 & 7.2 & 32.4 \\ \hline
$Attn^{S-M}$ & 53.5 & 34.8 & 7.9 & 34.1  \\ \hline
$Attn^{M-S}$ & 53.7 & 34.6 & 7.8 & 34.0 \\ \hline
$Attn^{S-S}$ + $Attn^{M-M}$  & 53.9 & 34.9 & 8.1 & 34.3\\ \hline 
$Attn^{S-S}$ + $Attn^{M-M}$ +  & \textbf{58.1} & \textbf{39.5} & \textbf{9.1} & \textbf{39.0} \\
$Attn^{S-M}$ + $Attn^{M-S}$ & & & \\ \hline 
\end{tabular}
\end{table}

\begin{figure*}[!htbp]
	\centering 
		\includegraphics[height=0.42\textheight]{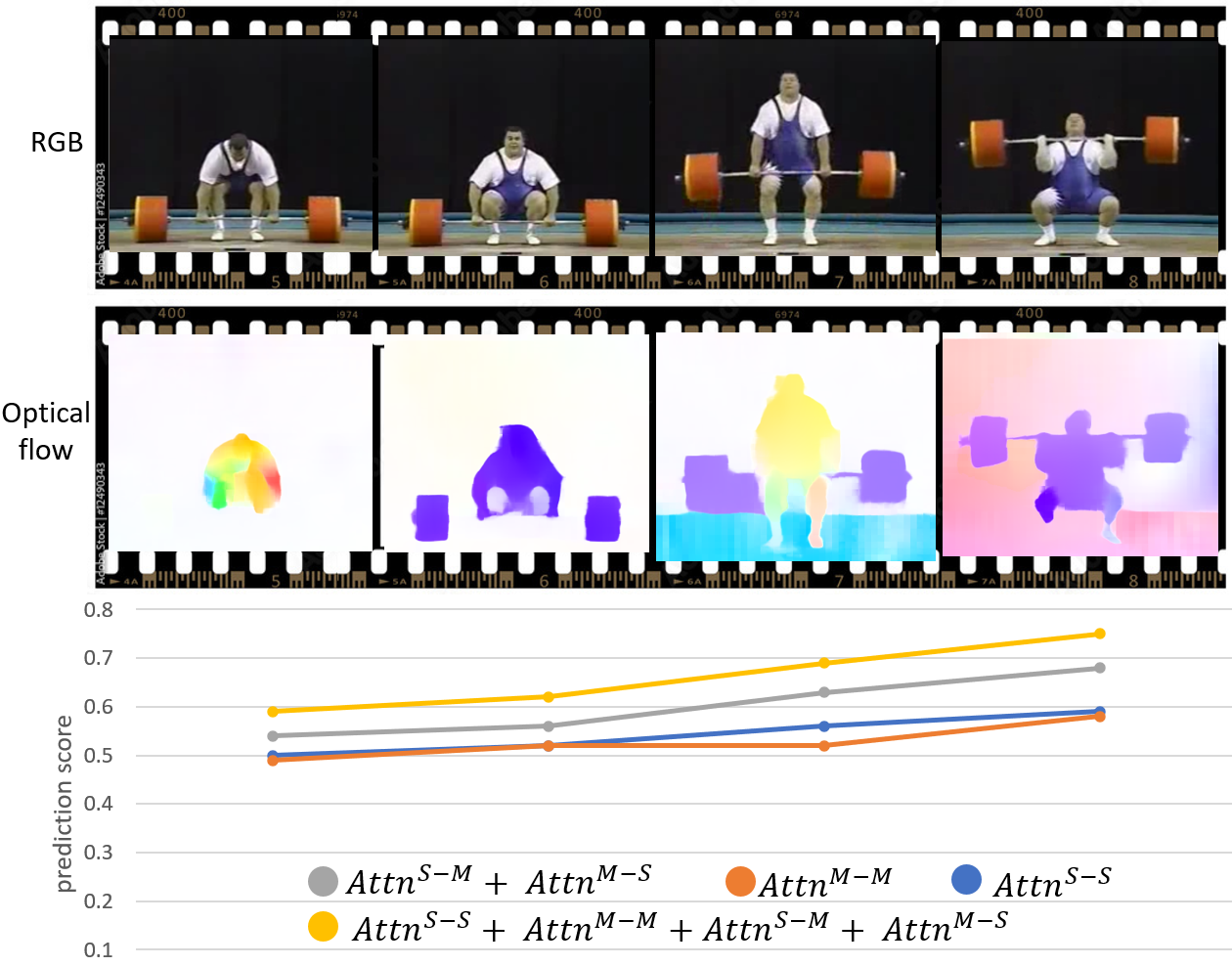} \\
	\caption{\textcolor{black}{Action prediction scores based on different multi-modal attention designs and time frames for the ground truth class. The action sample is selected from the validation set of the TUUMOS14 dataset. The sample video id is ``video\_validation\_0000151'' and the ground truth action class is ``Clean and Jerk''.} ~\label{Fig:vis1}}
\end{figure*}

\begin{figure*}[!htbp]
	\centering 
		\includegraphics[height=0.42\textheight]{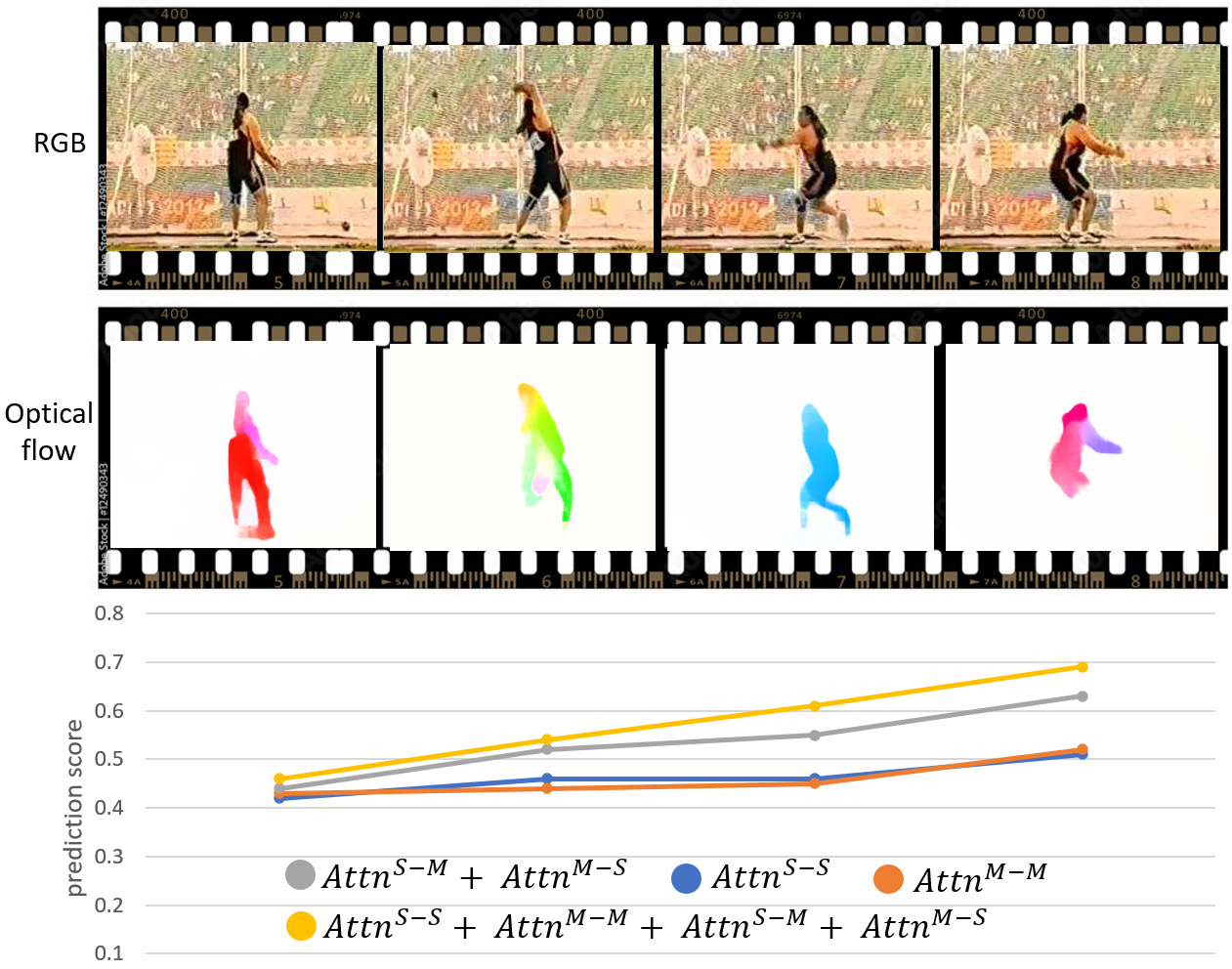} \\
	\caption{\textcolor{black}{Action prediction scores for different multi-modal attention designs and time frames. The action sample is selected from the validation set of the TUUMOS14 dataset. The sample video id is ``video\_validation\_0000311'' and the ground truth action class is ``Hammer Throw''.} ~\label{Fig:vis2}}
\end{figure*}

\textcolor{black}{Fig. \ref{Fig:vis1} and Fig. \ref{Fig:vis2} visualize the performance of different multi-modal attention designs based on two video samples selected from the validation set of the THUMOS14 dataset. Specifically, we showed the action prediction scores for the ground truth action classes based on various attention types and their combinations at different time frames. Here, higher prediction values mean higher reliability in detecting the correct action classes.  Investigating Fig. \ref{Fig:vis1} and Fig. \ref{Fig:vis2} closely shows that the highest prediction values are obtained when all the attention types are used ($Attn^{S-S} + Attn^{M-M} + Attn^{S-M} + Attn^{M-S}$), followed by the cross-modality attention ($Attn^{S-M} + Attn^{M-S}$).}

\textcolor{black}{Table 6 compares our multi-modal model with a standard model considering the impact of the motion distortion correction algorithm. The results show the overall action detection performance based on the THUMOS14 dataset. The standard model is defined by concatenating motion, and RGB features a common strategy when RGB and optical flow fields are used jointly.  The concatenated features with no motion correction algorithm define the \emph{Baseline} model that achieved the lowest average performance mAP of  63.9\%. The motion distortion correction algorithm enhanced both standard  (concatenated features) and our multi-modal model with an average mAP difference of 2.1\% and 1.9\%, respectively. We also showed the aforementioned results for the ActivityNet dataset in Table 7. Here, the baseline led to the least performance with an average mAP of 32.8\%. On the contrary,  the full model, including the multi-modal model and the motion distortion correction algorithm, resulted in the highest performance with an average mAP of 39\%. }

\begin{figure*}[!htbp]
\renewcommand{\tabcolsep}{0.5pt}
	\centering 
		\begin{tabular}{cc}
  \includegraphics[height=0.22\textheight]{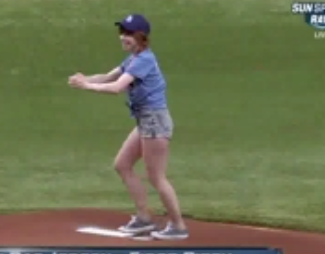} &
  \includegraphics[height=0.22\textheight]{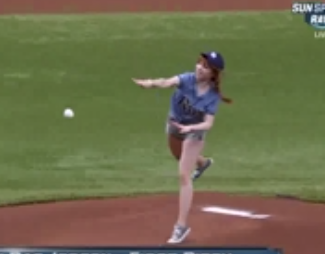} \\
    frame $t$ & frame $t + \tau$ \\
  \includegraphics[height=0.22\textheight]{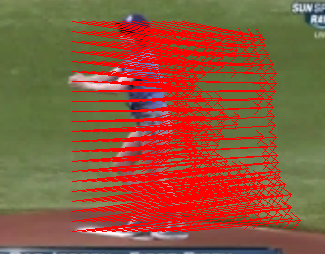} &
  \includegraphics[height=0.22\textheight]{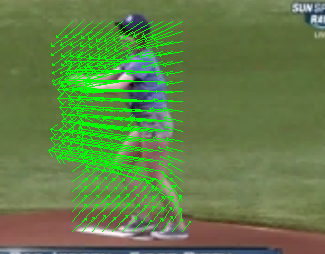} \\

  distorted motion vectors & our corrected motion  vectors \\ 

    	\end{tabular}
	\caption{An example of our motion distortion correction algorithm tested on an action sample from the THUMOS14 dataset \cite{idrees2017thumos}. We showed two consecutive sampled frames of $t$ and frame $t + \tau$, and consequently, the distorted motion vectors (in red) and our correction motion vectors (in green). In this example, the softball player is running to the left, while the camera also moves to the left more rapidly than the player. This causes a spatial-temporal inconsistency leading to incorrect extraction of motion vectors. Our motion distortion correction algorithm, however, fixes this issue. \label{Fig:exam_thu}}
\end{figure*}

We also show an example of our motion distortion correction algorithm evaluated on a sample from the THUMOS14 \cite{idrees2017thumos} dataset in Fig. \ref{Fig:exam_thu}. Here, both the softball player and camera are moving to the left, which causes a spatial-temporal inconsistency leading to motion distortion (red vectors) that is fixed by our motion distortion algorithm (green vectors).

\begin{table}[h!tbp]
	\centering
	\caption{\textcolor{black}{Comparison of the proposed multi-modal network with a standard modality concatenation approach with and without using our motion distortion correction algorithm on the overall action detection performance on the THUMOS14 dataset. The baseline model does not include the cross-modality attention mechanism and the motion distortion correction.}    } \label{Tab:motion_abl} 
\begin{tabular}{ >{\color{black}}c >{\color{black}}c >{\color{black}}c >{\color{black}}c >{\color{black}}c}
\hline
Option & maP@0.3 &  maP@0.5 & maP@0.7 & Avg \\ \hline
Concatenation without motion  & 81.2 &  69.0 & 40.8 & 63.9 \\ 
distortion correction (\textbf{Baseline}) & & & &  \\ \hline
Concatenation with motion  & 83.4 &  71.7 & 43.6 & 66.0 \\ 
distortion correction & & & &\\ \hline
Multi-modal without motion  & 83.5 &  71.9 & 44.2 & 66.6 \\ 
distortion correction & & & &  \\ \hline
Multi-modal with motion  & 85.2 &  73.4 & 45.3 & 68.5 \\ 
distortion correction (\textbf{Full model}) & & &  &\\ \hline
\end{tabular}
\end{table}

\begin{table}[h!tbp]
	\centering
	\caption{\textcolor{black}{Comparison of the proposed multi-modal network with a standard modality concatenation approach with and without using our motion distortion correction algorithm on the overall action detection performance on the ActivityNet dataset. The baseline model does not include  the cross-modality attention mechanism and the motion distortion correction.} } \label{Tab:motion_ab_actl} 
\begin{tabular}{ >{\color{black}}c >{\color{black}}c >{\color{black}}c >{\color{black}}c >{\color{black}}c}
\hline
Option & maP@0.5 &  maP@0.75 & maP@0.95 & Avg \\ \hline
Concatenation without motion  & 51.9 &  33.0 & 7.8 & 32.8 \\ 
distortion correction (\textbf{Baseline}) & & & & \\ \hline
Concatenation with motion  & 53.8 &  34.7 & 8.0 & 34.0 \\ 
distortion correction & & & &\\ \hline
Multi-modal without motion  & 56.0 &  37.9 & 8.6 & 37.5 \\ 
distortion correction & & & &  \\ \hline
Multi-modal with motion  & 58.1 &  39.5 & 9.1 & 39.0 \\ 
distortion correction (\textbf{Full model}) & & & &\\ \hline
\end{tabular}
\end{table}

Table \ref{Tab:time_anal} shows the efficiency analysis of different modules, which is based on running our algorithm on 10 seconds of an action video sample.

\begin{table}[h!tbp]
	\centering
	\caption{\textcolor{black}{Efficiency analysis of different modules} } \label{Tab:time_anal} 
\begin{tabular}{ >{\color{black}}c >{\color{black}}c >{\color{black}}c >{\color{black}}c >{\color{black}}c}
\hline
Module &   time (ms) \\ \hline
Motion distortion correction &  126 \\ \hline
$Attn^{S-S}$ & 640\\ \hline
$Attn^{M-M}$ &  645 \\ \hline
$Attn^{S-M}$ + $Attn^{M-S}$  & 762 \\ \hline 
$Attn^{S-S}$ + $Attn^{M-M}$ +  $Attn^{S-M}$ + $Attn^{M-S}$ & 872\\ 
 \hline
\end{tabular}
\end{table}

\subsection{Experimental results on instructional activity dataset}

\begin{figure*}[!htbp]
	\centering 
		\frame{\includegraphics[height=0.4\textheight]{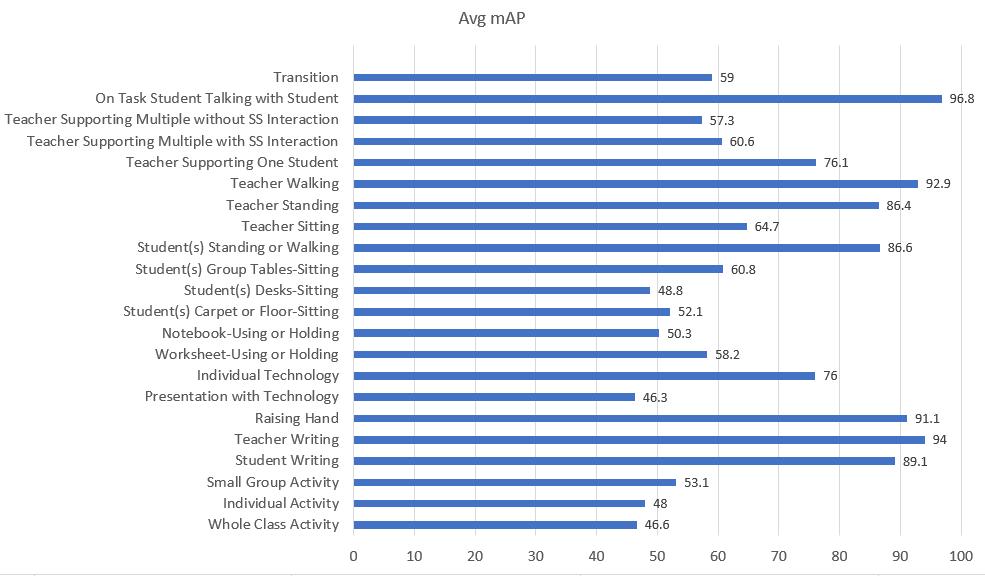}} \\
	\caption{Average performance of our proposed method per action class evaluated on our instructional activity dataset.  ~\label{Fig:results_per_class}}
\end{figure*}

Table \ref{Tab:results_ours} shows the comparative results on our instructional activity dataset based on the average mAP performance. We compared our method with  SE \cite{piergiovanni2018CVPR} (CVPR 2018), GA \cite{shi2020CVPR} (CVPR 2020), LST \cite{xu2021NIPS} (NIPS 2021), MLAD \cite{tirupattur2021CVPR} (CVPR 2021), BF \cite{lin2021CVPR} (CVPR 2021), and COLA \cite{zhang2021CVPR} (CVPR 2021). 
Fig. \ref{Fig:results_per_class} indicates the average mAP performance of our proposed method on our instructional activity dataset separated for each class label. 

\textcolor{black}{Our proposed method outperformed the best other results (MLAD) with an average mAP difference  of 22.9\%.}
\textcolor{black}{Our method performed significantly better than others because of the following reasons: (1) with a high density of people in the instructional videos, people may have a variety of complex motions. So, our multi-modal attention mechanism can better model the classroom videos by capturing the correlations between two spatial and motion modalities rather than modeling these modalities separately. (2) With this high human density, the motion feature representation can be highly impacted by camera movement, which is handled by our motion distortion correction algorithm.}

\begin{table}[h!tbp]
	\centering
	\caption{Comparison of our proposed method with the state-of-the-art approaches on our instructional activity dataset.} \label{Tab:results_ours} 
\resizebox{\textwidth}{!}{\begin{tabular}{cccccccc}
\hline
Method & MLAD \cite{tirupattur2021CVPR} & SE \cite{piergiovanni2018CVPR} & BF \cite{lin2021CVPR} & GA \cite{shi2020CVPR} & LST \cite{xu2021NIPS} & COLA \cite{zhang2021CVPR} & \textbf{Ours (MMNet)}\\ \hline
Avg & 45.2 & 36.7 & 25.3 & 26.0 & 42.0 & 34.1 & 68.1
\\ \hline
\end{tabular}}
\end{table}

\section{Conclusions}
This paper proposes a novel transformer network for detecting actions in untrimmed videos. Our transformer network utilizes a new multi-modal attention mechanism to capture the correlative patterns between spatial (RGB) and motion (optical flow) features. Such correlative features improve the expressive power of action features. To employ the motion (optical flow) inputs more effectively, we put forth a motion distortion correction algorithm to handle camera movements that can severely distort the motion vectors represented in the optical flow fields.
In terms of the empirical aspect of the work, we introduce a new instructional activity dataset captured from elementary schools.
Our proposed method outperformed the state-of-the-art approaches on two public benchmarks, THUMOS14 and ActivityNet, as well as on our newly created instructional activity dataset. 

Our study is beneficial for other researchers in the field as we are the first to suggest capturing the correlative patterns between RGB and optical flow using an effective multi-modal attention mechanism.  Moreover, our novel motion distortion correction algorithm is advantageous in dealing with camera movement, which is common in real-world scenarios and in the wild and should have broad applicability. 

\textbf{Future work}
While our motion distortion algorithm is effective in dealing with camera movements, it still depends on a person detection algorithm to segment the background and foreground. We suggest modeling the background, preferably within the action detection network itself, as a future direction. 
Moreover, for our multi-modal transformer, we suggest separating the semantics (for both RGB and optical flow) in the scene to capture the correlative patterns among local objects/subjects instead of whole-action frames.

\textbf{Acknowledgements}

This material is based upon work supported by the National Science Foundation under Grant No. 2000487 and the Robertson Foundation under Grant No. 9909875. Any opinions, findings, conclusions, or recommendations expressed in this material are those of the authors and do not necessarily reflect the views of the funding organizations. The authors would like to thank Mr. Tyler Spears for his help in editing.

\bibliography{mybibfile}

\noindent\parbox{11.3cm}{\parpic{\includegraphics[width=30mm,scale=0.1]{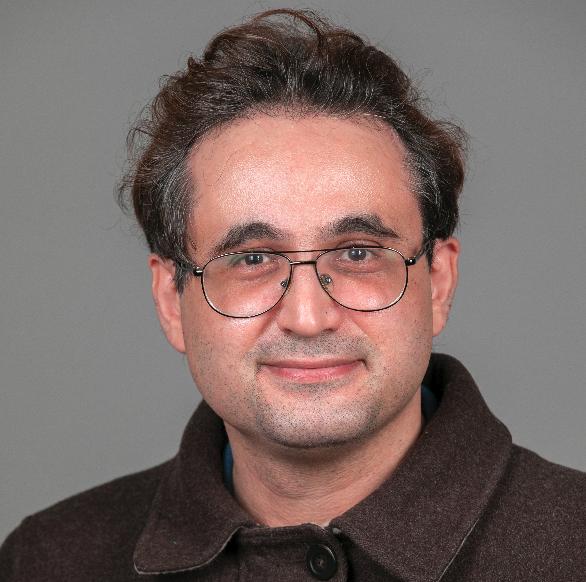}}{\small\quad {\bf Matthew {Korban}} received his BSc and MSc degree in Electrical Engineering in 2013 from the University of Guilan, where he worked on sign language recognition in video. He received his PhD in Computer Engineering from Louisiana State University. He is currently a Postdoc Research Associate at the University of Virginia, working with Prof. Scott T. Acton. His research interest includes Human Action Recognition, Early Action Recognition, Motion Synthesis, and Human Geometric Modeling in Virtual Reality environments.}\\[1mm]}

\noindent\parbox{11.3cm}{\parpic{\includegraphics[width=30mm,scale=0.1]{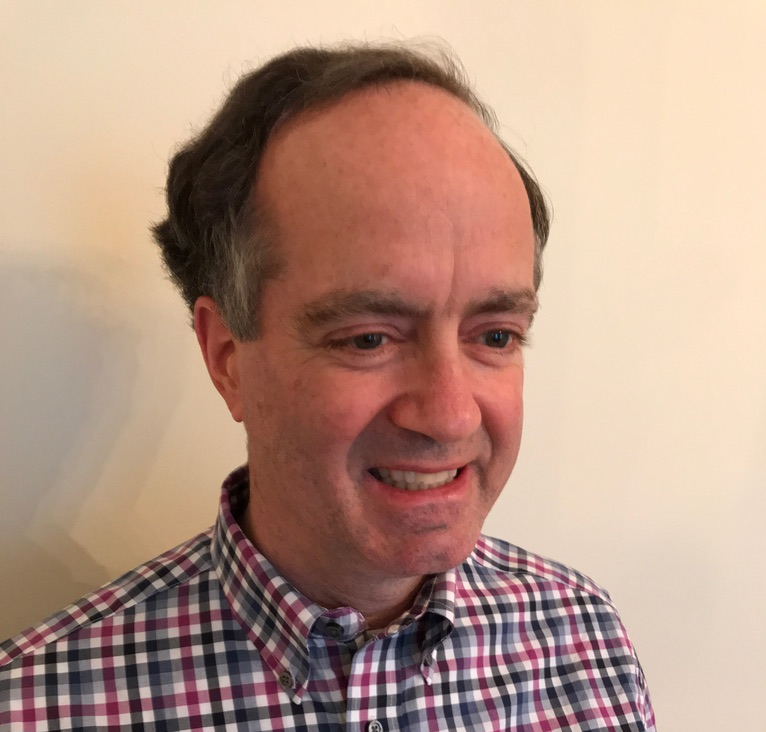}}{\small\quad {\bf Peter {Youngs}} is a professor in the Department of Curriculum, Instruction and Special Education at the University of Virginia. He studies how neural networks can be used to automatically classify instructional activities in videos of elementary mathematics and reading instruction. He currently serves as co-editor of American Educational Research Journal. }\\[1mm]}

\noindent\parbox{11.3cm}{\parpic{\includegraphics[width=30mm,scale=0.1]{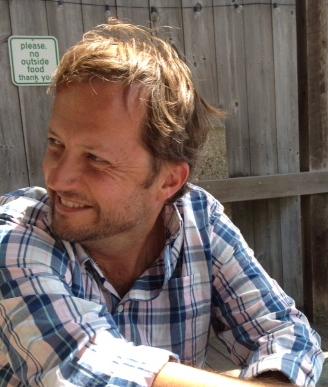}}{\small\quad {\bf Scott T. {Acton}} received his M.S. and Ph.D. degrees at the University of Texas at Austin. He received his B.S. degree at Virginia Tech. Professor Acton is a Fellow of the IEEE ``for contributions to biomedical image analysis.”
Currently, Acton is a Professor and Chair of Electrical and Computer Engineering at the University of Virginia. He recently finished three years as Program Director in the National Science Foundation. Professor Acton’s laboratory at UVA is called VIVA - Virginia Image and Video Analysis. They specialize in bioimage analysis problems.  Professor Acton has over 300 publications in the image analysis area including the books \textit{Biomedical Image Analysis: Tracking} and \textit{Biomedical Image Analysis: Segmentation}. He was the 2018 Co-Chair of the IEEE International Symposium on Biomedical Imaging. Professor Acton was recently Editor-in-Chief of the IEEE Transactions on Image Processing (2014-2018).}\\[1mm]}

\end{document}